\documentclass{article} 
\usepackage{iclr2026_conference,times}

\usepackage{amsmath,amsfonts,bm}

\def\eqref#1{equation~\ref{#1}}

\def\1{\bm{1}}

\DeclareMathAlphabet{\mathsfit}{\encodingdefault}{\sfdefault}{m}{sl}
\SetMathAlphabet{\mathsfit}{bold}{\encodingdefault}{\sfdefault}{bx}{n}

\usepackage{hyperref}
\usepackage{url}
\usepackage{graphicx}
\usepackage{mathtools}
\usepackage{amsthm}
\usepackage{float}
\usepackage{subcaption}
\usepackage{multirow}
\usepackage{booktabs}
\usepackage{bm}

\title{Investigating intra-abstraction policies for non-exact abstraction algorithms}

\author{Robin Schmöcker \\
Institute for Information Processing\\
Leibniz University Hannover\\
Hannover, Germany \\
\texttt{schmoecker@tnt.uni-hannover.de} \\
\And
Alexander Dockhorn \\
SDU Metaverse Lab \\
University of Southern Denmark \\
Odense, Denmark \\
\texttt{adoc@mmmi.sdu.dk} \\
\And
Bodo Rosenhahn \\
Institute for Information Processing\\
Leibniz University Hannover\\
Hannover, Germany \\
\texttt{rosenhahn@tnt.uni-hannover.de} \\
}

\iclrfinalcopy 
\begin{document}

\maketitle

\begin{abstract}
One weakness of Monte Carlo Tree Search (MCTS) is its sample efficiency which can be addressed by building and using state and/or action abstractions in parallel to the tree search such that information can be shared among nodes of the same layer. The primary usage of abstractions for MCTS is to enhance the Upper Confidence Bound (UCB) value during the tree policy by aggregating visits and returns of an abstract node. However, this direct usage of abstractions does not take the case into account where multiple actions with the same parent might be in the same abstract node, as these would then all have the same UCB value, thus requiring a tiebreak rule. In state-of-the-art abstraction algorithms such as pruned On the Go Abstractions (pruned OGA), this case has not been noticed, and a random tiebreak rule was implicitly chosen. In this paper, we propose and empirically evaluate several alternative intra-abstraction policies, several of which outperform the random policy across a majority of environments and parameter settings.
\end{abstract}

\section{Introduction}
\label{sec:intro}
\noindent Monte Carlo Tree Search \citep{BrownePWLCRTPSC12,coulom06} (MCTS) is a domain-independent and on-the-fly applicable search algorithm which is the state-of-the-art for some decision-making tasks such as Mahjong \citep{mahjongIS}. Even though machine learning (ML) approaches outcompete standard MCTS in many applications such as AlphaGo for Go \citep{SilverHMGSDSAPL16} or OpenAI Five for Dota2 \cite{dota2openaifive}, ML techniques require a resource intensive training phase which makes it undesirable for some domains. For example, in game development, ML agents are rarely used since they would have to be costly retrained every time the game is updated. 
Even though this is not within the scope of this paper, foundational improvements to MCTS could also potentially translate to improvements in ML algorithms utilizing this search such as Alpha Zero \citep{alphazero}. Hence, it is still worth dedicating research effort to MCTS.

One research area to improve MCTS is using abstractions that aim at reducing the search space by grouping states and actions in the current MCTS search tree to enable an intra-layer information flow \citep{uctJiang,AnandGMS15,OGAUCT}, by averaging the visits and returns of all abstract action nodes in the same abstract node used for the Upper Confidence Bounds (UCB) formula in the tree policy.
Inevitably, there are action nodes with the same parent state and same abstract node, which results in multiple actions having the exact same UCB value during the tree policy. Without giving this case special treatment, state-of-the-art algorithms like pruned On the Go Abstractions (pruned OGA) \citep{OGAUCT} simply perform tiebreaking exactly as in the non-abstracted case. In the case of pruned OGA, this is done randomly. 

In this paper, we aim to tackle exactly this problem by proposing and evaluating several random-policy alternatives, several of which significantly enhance OGA's performance across a variety of environments and parameter settings.
The contributions of this paper can be summarized as follows:
\begin{itemize}

    \item We propose the \textbf{Alternating State And State-Action Pair Abstractions (ASASAP)} framework, which generalizes the abstractions built by most MCTS-based abstraction algorithms, including Automatic State abstractions (AS), Abstractions of State-Action Pairs (ASAP), and OGA.

    \item We empirically show that the case of having multiple abstracted actions with the same parent is not an edge case and occurs frequently.

    \item We propose and evaluate seven intra-abstraction policies as alternatives to the random policy, namely: \textbf{UCT}, \textbf{FIRST}, \textbf{GREEDY}, \textbf{MOST\_VISITS}, \textbf{LEAST\_VISITS}, \textbf{LEAST\_OUTCOMES}, and \textbf{RANDOM\_GREEDY}. The former, UCT, performs best overall, is a parameter-free drop-in improvement to OGA, as it performs either equally well or better across a wide range of parameters and problem settings. Furthermore, it causes only a negligible runtime overhead (see Tab.~\ref{tab:runtimes}). 

\end{itemize}

The paper is structured as follows. Firstly, in \textbf{Section} \ref{sec:foundations}, we lay the theoretical groundwork for this paper. In particular, we define the ASASAP framework, which helps us introduce and classify other abstraction frameworks such as ASAP or AS from the literature. Next, in \textbf{Section} \ref{sec:method}, we reiterate the intra-abstraction policy problem, describe seven alternatives, and illustrate on a concrete game tree how one of these modifications, using UCT as the intra-abstraction policy, can provably improve the performance. We then describe our experiment setup in \textbf{Section} \ref{sec:experiment_setup}. The experimental results are presented and discussed in \textbf{Section} \ref{sec:experiments}, where we first measure the number of times an intra-abstraction policy has to be queried in the first place, followed by a thorough analysis of all proposed intra-abstraction policies with a focus on UCT. At the end, in \textbf{Section} \ref{sec:future_work}, we briefly summarise our findings and provide an outlook for future work. 

\section{Foundations of non-learned domain-independent abstractions}
\label{sec:foundations}
\noindent \textbf{Problem model and optimization objective:}
We use finite MDPs \citep{sutton2018reinforcement} as the model for sequential, perfect-information decision-making tasks. Here, $\Delta(X)$ denotes the probability simplex of a finite, non-empty set $X$ and $\mathcal{P}(X)$ denotes the power set of $X$.

\textit{Definition:}
    An \textit{MDP} is a 6-tuple $(S,\mu_0,\mathbb{A},\mathbb{P}, R, T)$ where the components are as follows:
    \begin{itemize}
        \item $S \neq \emptyset$ is the finite set of states, $T \subseteq S$ is the (possibly empty) set of terminal states, and  $\mu_0 \in \Delta(S)$ is the probability distribution for the initial state.
        \item $\mathbb{A}\colon S \mapsto A$ maps each state $s$ to the available actions $\emptyset \neq \mathbb{A}(s) \subseteq A$ at state $s$ where $|A| < \infty$.
        \item $\mathbb{P}\colon S \times A \mapsto \Delta(S )$ is the stochastic transition function where we use $\mathbb{P}(s^{\prime} |\: s,a)$ to denote the probability of transitioning from $s \in S$ to $s^{\prime} \in S$ after taking action $a \in \mathbb{A}(s)$ in $s$.
        \item $R \colon S \times A \mapsto \mathbb{R}$ is the reward function.
    \end{itemize}

\noindent From hereon, let $M = (S,\mu_0,\mathbb{A},\mathbb{P}, R, T)$ be an MDP. Using the same notation as \cite{AnandGMS15}, we also define
    \mbox{$P \coloneqq \{(s,a)\: | \: s \in S, a \in \mathbb{A}(s)\}$}
as the set of all state-action pairs. The goal is to find an agent $\pi$ that we model as a mapping from states to action distributions
    $\pi \colon S \mapsto \Delta(A)$
such that $\pi$ maximizes the expected episode's return, where the (discounted) return for of episode $s_0,a_0,r_0, \dots, s_n,a_n,r_n,s_{n+1}$ with $s_{n+1} \in T$ is given by $\gamma^0 r_0 + \ldots + \gamma^n r_n$.

\noindent \textbf{Abstraction frameworks}
Next, we will define a general abstraction framework that includes most of the here-presented abstraction algorithms and captures their core working principle. We bluntly call this framework \textbf{A}lternating \textbf{S}tate \textbf{A}nd \textbf{S}tate-\textbf{A}ction-\textbf{P}air \textbf{A}bstractions (ASASAP) whose purpose is to unify parts of the abstraction zoo.
The idea of ASASAP is to alternately construct a state abstraction from a state-action-pair abstraction and vice versa. For our purposes, we simply define state and action abstractions as equivalence relations (equivalently partitions) of the state or action space. In the supplementary materials in Section \ref{sec:example_asap}, we show a concrete example of how an ASAP abstraction (a special case of ASASAP) is built.

\textit{Definition}:
    We call the equivalence relation  \mbox{$\mathcal{H} \subseteq P \times P$} induced by some $n \in \mathbb{N}$, 
    some initial state equivalence relation $\mathcal{E}_0 \subseteq S \times S$, mappings \mbox{$f \colon \mathcal{P}( S \times S) \mapsto \mathcal{P}(P \times P)$} and \mbox{$g\colon \mathcal{P}(P \times P) \mapsto \mathcal{P} (S \times S)$} to equivalence relations an $ASASAP_{f,g,n,\mathcal{E}_0}$
    abstraction if it is of the form
    \begin{align}
    \mathcal{H} &= \mathcal{H}_n, \\
    \mathcal{H}_{i+1} &= f(\mathcal{E}_i) & \forall i, \\
    \mathcal{E}_{i+1} &= g(\mathcal{H}_{i+1}) & \forall i.
    \end{align}
If additionally $\mathcal{H}$ is invariant to any number of additional applications of $f$ and $g$, then we call it \textit{converged}.

Next, we will present some concrete instances of ASASAP from the literature. Firstly, \cite{uctJiang} used- and \cite{GivanDG03} proposed AS-UCT (the name was given by \cite{AnandGMS15}), which defines $g_{\text{AS}}(\mathcal{H}_{i+1})$ as grouping states if and only if they have identical legal actions and they are pairwise equivalent:
\begin{equation}
    \begin{aligned}
   & (s_1,s_2) \in g(\mathcal{H}_{i+1})  \iff \mathbb{A}(s_1) = \mathbb{A}(s_2) \: \land \: \\
        &\forall a_1 \in \mathbb{A}(s_1):\: 
        ((s_1,a_1),(s_2,a_1)) \in \mathcal{H}_{i+1}.
    \end{aligned}
\end{equation}

And any state-action-pair $(s_1,a_1),(s_2,a_2)$ is equivalent i.e. $((s_1,a_1),(s_2,a_2)) \in f_{AS}(\mathcal{E}_{i})$ if and only if the state-action pairs have similar immediate rewards and transition distributions:
\begin{equation}
    \begin{aligned}
         \quad | R(s_1,a_1) - R(s_2,a_2) | 
        & \leq \varepsilon_{\text{a}} \\
        \quad \textrm{and } F \coloneqq \sum \limits_{x \in \mathcal{X}} \bigg| \sum \limits_{s^{\prime} \in x} 
        \mathbb{P}(s^{\prime}|\: s_1,a_1) - \mathbb{P}(s^{\prime}|\: s_2,a_2) \bigg| 
       &  \leq \varepsilon_{\text{t}},
    \end{aligned}
\end{equation}
where $\mathcal{X}$ are the equivalence classes of $\mathcal{E}_{i}$ and $\varepsilon_{\text{t}}, \varepsilon_{\text{a}} \geq 0$. In general, for $\varepsilon_{\text{t}},\varepsilon_{\text{a}} > 0$, $f_{AS}(\mathcal{E}_i)$ is not an equivalence relation because transitivity is not guaranteed. Hence, any abstraction algorithms using these need to slightly modify $f_{\text{AS}}(\mathcal{E}_i)$ to obtain an equivalence relation.
The reason for allowing $\varepsilon_{\text{a}}$ and $\varepsilon_{\text{t}}$ to be greater than $0$, is to find more correct abstractions at the cost of potentially abstracting state-action-pairs or states that do not have the same value. The experiments of this paper confirm that this can be beneficial.

To allow for the detection of more symmetries, \cite{AnandGMS15} proposed ASAP abstractions that are based on \cite{ravindran2004approximate} homomorphism condition that does not require there to be a 1-to-1 action match but only a mapping of actions to each other, concretely $g_{\text{ASAP}}(\mathcal{H}_{i+1})$ is defined as

\begin{equation}
    \begin{aligned}
   & (s_1,s_2) \in g_{\text{ASAP}}(\mathcal{H}_{i+1})  \iff \\
        &\forall a_1 \in \mathbb{A}(s_1) \, \exists a_2 \in \mathbb{A}(s_2):  
        ((s_1,a_1),(s_2,a_2)) \in \mathcal{H}_{i+1} \\
        &\forall a_2 \in \mathbb{A}(s_2) \, \exists a_1 \in \mathbb{A}(s_1):  
        ((s_1,a_1),(s_2,a_2)) \in \mathcal{H}_{i+1}.
    \end{aligned}
\end{equation}
The action abstraction $f_{\text{ASAP}}$ is the same as the previously defined $f_{AS}$ using $\varepsilon_{\text{t}}=\varepsilon_e=0$, however, as we will later see there is nothing that would prevent us from choosing epsilon values greater than zero here.

\noindent \textbf{Abstractions for search: }
Constructing ASAP or AS abstractions until convergence for an entire MDP is oftentimes infeasible, and such a computation would significantly hamper the runtime. Hence, ASAP-UCT \citep{AnandGMS15}, AS-UCT \citep{AnandGMS15,uctJiang}, and OGA-UCT \citep{OGAUCT} build their ASASAP abstraction on the \textbf{local-layered MDP} rooted at the state $s_d$ where the decision has to be made.

\textit{Definition:}
    The state space of the \textit{layered MDP} of $M$ is $S \times \{0,\dots,h\}$ where $h \in \mathbb{N}$ is the horizon and if $(s,n)$ is a successor state of $(s^{\prime},n^{\prime})$, then $n = n^{\prime} +1$ and any initial state has $n=0$. Additional terminal states are $S \times \{h\}$. The \textit{local-layered MDP} rooted at $s_d$ is the layered MDP of $M$ but with its states, actions, and possible state-action-pair-successors restricted to those present in the current search graph.

In local-layered MDPs, a converged ASAP or AS abstraction can be efficiently computed with dynamic programming, where one requires only the abstraction of the previous layer to compute the abstractions for the next. In ASAP-UCT and AS-UCT, an ASAP/AS-like abstraction is built in regular intervals on the current MCTS (for details on MCTS, see Section \ref{sec:mcts}) search graph using an
initial state equivalence relation that groups all terminal states of the same layer, groups all non-fully-expanded nodes of the same layer, and puts all remaining nodes in their own abstract node of size one. The abstraction built by ASAP/AS-UCT differs only from the ASAP/AS abstraction in that non-fully-expanded nodes are never grouped with fully-expanded nodes. This non-grouping condition also applies to OGA \citep{OGAUCT}. For the later experiments, we will also experiment with grouping partially explored state nodes as in ASAP-UCT, but for OGA. We refer to this parameter as $\text{PG} \in \{0,1\}$ where $0$ refers to no partial grouping.

Unlike ASAP-UCT and AS-UCT, the successor of ASAP-UCT, OGA-UCT, does not compute its respective abstraction from the ground up but rather attempts to approximate the ASAP abstraction by rebuilding only parts of its current abstraction. More concretely, OGA-UCT tests every $K$-th Q node visit if the abstraction needs to be updated (e.g., new successors were sampled that invalidate a previous abstraction). If so, the parent's abstraction is recursively updated too.

A core weakness of ASAP abstractions is their exactness, which causes them to not be able to deal with stochasticity well. Hence, \cite{OGAUCT} directly proposed \textit{pruned OGA} as an improvement to OGA-UCT, which is the same as OGA-UCT except that for the abstraction construction step for each state-action pair with $n$ successors with respective probabilities $p_1,\dots,p_n$ only those with $p_i > \alpha \cdot \max \{p_1,\dots,p_n\},\ \alpha \in [0,1]$ are considered. Also in this paper, we consider $(\varepsilon_{\text{a}},\varepsilon_{\text{t}})$-OGA \citep{ogacad} which is equivalent to OGA-UCT except that one allows for $\varepsilon_{\text{a}},\varepsilon_{\text{t}}$ to be greater than $0$. Since this does not induce an equivalence relation, the abstraction construction process has to be slightly modified as detailed by \cite{ogacad}.

\noindent \textbf{Abstraction usage:}
Thus far, we have only discussed how to build abstractions but not how to use them. The key mechanism that all here-presented MCTS-based abstraction methods use (e.g. AS-UCT, ASAP-UCT, OGA-UCT) is only to modify the tree policy by enhancing the UCB value. The UCB value for an action is enhanced by using the aggregated visits and returns of all actions that are part of the same abstract action (i.e. equivalence class). In particular, state abstractions are not used at this stage. These are only needed as an intermediate step to find action abstractions. 
Only AS-UCT differs slightly from this approach as it only aggregates the statistics of actions that additionally have the same abstract parent. This is because AS-UCT was originally intended as a state only abstraction which is why it did not decouple action and state abstractions.

The intra-abstraction policies that we will later propose only affect the abstraction usage component of an abstraction algorithm. They do \textbf{not} modify the abstraction-building process itself.

\noindent \textbf{Other automatic abstraction algorithms:}
A different abstraction paradigm is employed in PARSS by \cite{HostetlerFD15} in which instead of iteratively building larger and larger abstractions, one starts by grouping all successors of each state-action pair and then refining these groups in parallel to the search. Similarly, AUPO by \cite{aupo} initially groups all root node state-action pairs and only splits two state-action pairs if there is statistical evidence that they have different $Q^*$ values.
While not in scope of this paper, there exists abstraction techniques for imperfect information and/or continuous problem domains \citep{HoergerKKY24} as well as ML-based approaches which learn and plan on abstractions of the original problem \citep{OzairLRAOV21,KwakHKLZ24,ChitnisSKKL20}.
Research effort has also been dedicated towards automatic abstractions of the transition function, which on an abstract level can be described as pruning certain successors from the transition function \citep{SokotaHAK21,YoonFGK08,YoonFG07,saisubramanian2017optimizing}. A comprehensive survey on abstraction algorithms is provided by \cite{mysurvey}.

\section{Method}
\label{sec:method}
\noindent \textbf{Intra-abstraction policies:}
A consequence of ASAP's key idea to decouple state and action abstractions is that two state-action pairs may be abstracted even when they have the same parent. This, however, leads to the thus-far overlooked problem that any two abstract Q nodes with the same parent will have an identical UCB value (see Section \ref{sec:mcts}) as they have the same number of aggregated visits and returns. Hence, a tiebreaking rule is needed, which we refer to as an  \textbf {intra-abstraction policy}. \cite{OGAUCT} implicitly chose a random intra-abstraction policy. While this random policy causes no harm when the abstractions are lossless, when dealing with lossy abstractions (i.e., those where states or actions could be abstracted even when they do not have the same value under optimal play) a random policy can be detrimental to performance as we will show in the experiment section \ref{sec:experiments}.

We propose a number of alternative intra-abstraction policies to choose an action within the selected abstract node. The intra-abstraction policy can be split into two phases. One for the decision policy (i.e. for the final decision at the root node) and one for the tree policy.

We separate the to-be-proposed methods into four groups. The first group encompasses the implicitly used methods from the literature. The second group includes policies that focus on exploration, the third group focuses on exploitation, and the fourth group is a mix of both. 

\textbf{1. RANDOM}: Randomly choose an action with uniform probability for both the decision and tree policy. This is the one used by \cite{AnandGMS15} in ASAP-UCT.

\textbf{2. FIRST}: Simply choose the first encountered ground action for both the decision and tree policy. Though we do not believe that this method would perform well, this is one of the standard tiebreaking rules for MCTS and might be accidentally implicitly used in an abstraction algorithm.

\vspace{-1em}
\rule{\textwidth}{0.4pt}
\vspace{-1.2em}

\textbf{3. RANDOM\_GREEDY}: Choose the action randomly during the tree policy and greedily during the decision policy. We include this method as an ablation to pin down the influence of simply being greedy in the decision policy.
     
\textbf{4. LEAST\_VISITS}: Choose the action with the least number of visits with a random tiebreak if two actions have the same number of visits. Greedy policy as the decision policy. This policy is closest to RANDOM except that visits are distributed perfectly evenly.

\textbf{5. LEAST\_OUTCOMES}: Choose the action with the minimal probability sum of all thus-far sampled successors. Greedy policy as the decision policy. The idea is to allow the detection of any faulty abstractions as soon as possible.

\vspace{-1em}
\rule{\textwidth}{0.4pt}
\vspace{-1.2em}

\textbf{6. GREEDY}: Choose the action with the highest Q value.

\textbf{7. MOST\_VISITS}: Choose the action with the most visits. Greedy policy is the decision policy. The idea behind this policy is to increase the search depth, because when the abstraction does not change, the action that wins the first tiebreak will always be chosen, as it will be the only one receiving visits. However, this may come at the cost of exploration.

\vspace{-1em}
\rule{\textwidth}{0.4pt}
\vspace{-1.2em}

\textbf{8. UCT}: Choose the action whose ground visits and values maximize the UCB value (see Section \ref{sec:mcts}) using the same exploration constant as the UCB selection for the abstract action. Greedy policy is the decision policy.
    
All policies use a random tiebreak e.g., when two ground actions have the same number of visits when using the LEAST\_VISITS policy.

\begin{figure}
    \centering
\includegraphics[width=0.4\textwidth]{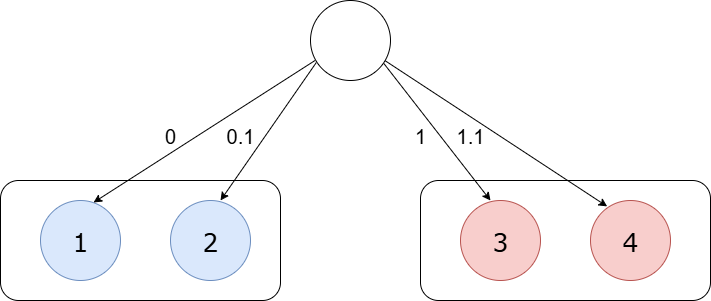}  
   \caption{Assume that MCTS with the visualized fixed action abstraction is run on the following deterministic depth-1 game tree. This abstraction would also be discovered by $(0.1,0)$-OGA since when all four actions are played at least $K$ times, $(0.1,0)$-OGA will have abstracted actions 1,2, and 3,4. While the visits will converge to choosing the abstract node with actions 3 and 4 for both the RANDOM and UCT intra-abstraction policy, RANDOM will distribute its visits uniformly amongst 3 and 4, resulting in an average payoff of $1.05$. This shows that with RANDOM, convergence to the optimal action is in general not guaranteed. In contrast, the UCT intra-abstraction policy guarantees convergence to the average payoff of $1.1$ by converging to action 4. 
   }
    \label{fig:abs_drop}
\end{figure}

\noindent \textbf{Case study RANDOM vs UCT: }
Next, we are going to study the theoretical properties of the RANDOM versus the UCT intra abstraction policy. Firstly, given arbitrary abstractions, no guarantees that go beyond those of any OGA-based methods can be made, as one can always construct abstractions such that intra-abstraction policies would never be queried. However, if we assume a special case of abstractions, which are those that only group state-action pairs with the same parent node, then guarantees can be made.
Firstly, RANDOM in combination with an arbitrary  same-parent state-action pair abstraction is not guaranteed to converge to the optimal action. An example where RANDOM fails to find the optimal action is given in Fig.~\ref{fig:abs_drop}. In contrast, using UCT will always converge to the optimal action. Concretely, assume a decision has to be made at state $s_{\text{d}}$. 

\textit{Theorem 1}: Let $\mathcal{E}$ be a same-parent state-action pair abstraction of the local-layered MDP rooted at $s_{\text{d}}$. Consider MCTS that uses the aggregated abstract visits and returns of $\mathcal{E}$ for the UCB value calculation for any state-action pair in combination with the UCT intra-abstraction policy. This MCTS version's ratio of root node visits of the optimal action(s) to the number of iterations will converge in probability to 1. 

The proof of this theorem is provided in the supplementary materials in Section \ref{sec:proof}. A direct consequence of this theorem is that pruned OGA's or $(\varepsilon_{\text{a}},\varepsilon_{\text{t}})$-OGA's root visits will also, in probability, converge to the optimal action(s) if they were slightly modified to only group state-action pairs with the same parent.
This is because eventually their abstractions will converge since all the MDP's state-action pairs will be visited and all their outcomes will be sampled almost surely, which allows only to apply Theorem 1.

\section{Experiment setup}
\label{sec:experiment_setup}
In this section, we describe the general experiment setup. Any deviations from this setup will be explicitly mentioned.

\textbf{Problem models:}
For this paper, we ran our experiments on a variety of MDPs, all of which are either from the International Probabilistic Planning Conference \citep{grzes2014ippc} or are commonly used in the abstraction algorithm literature \citep{AnandGMS15,OGAUCT,HostetlerFD15,YoonFGK08,uctJiang}. All models were chosen such that they are not simultaneously sparse reward and deterministic, as in that case any intra-abstraction policy for the here-considered abstraction algorithm would have no effect at all.
We ran all of our experiments on the finite-horizon versions of the considered MDPs with a default horizon of 50 steps and a discount factor $\gamma=1$. 
If the reader is not familiar with any of the domains we used for the experiments, we provide a brief description for each MDP in the supplementary materials in Section~\ref{sec:problem_descriptions}. 

\textbf{Parameters:}
Since the problem models vary in the scale of the encountered Q values, the dynamic exploration factor Global Std \citep{demcts} is used which
has the form $C \cdot \sigma$ where $\sigma$ is the standard deviation of the Q values of all nodes in the search tree and $C \in \mathbb{R}^+$ is some fixed parameter.
Furthermore, we always use $K=3$ as the recency counter, which was proposed by \cite{OGAUCT}.

\textbf{Evaluation:}
Each data point that we denote in the remaining sections of this paper (e.g. agent returns) is the average of at least 2000 runs. Whenever we denote a confidence interval for a data point, then this is always a confidence interval with a confidence level of 99\% which is $\approx 2.33$ times the standard error. Furthermore, we use a borda-like ranking system to quantify agents' performances; in particular, we use \textit{pairings} and \textit{relative improvement scores}. For details, see supplementary Section \ref{subsec:scors_defs}.

\textbf{Reproducibility:}
Our code was compiled with g++ version 13.1.0 using the -O3 flag (i.e. aggressive optimization) and it is publicly available \citep{repo}. 

\newpage
\section{Experiments}
\label{sec:experiments}
\noindent \textbf{Why intra-abstraction policies are needed:}
For the first set of experiments, we validate that the case where intra-abstraction policies are required, i.e., two actions with the same parent but the same abstract node, is not an edge case but occurs frequently. Tab.~\ref{tab:tiebreak_stats} in the supplementary materials lists these statistics, showing that even for the least coarse abstraction setting, i.e., $\varepsilon_{\text{t}}=\varepsilon_{\text{a}}=0$ and no partial grouping, there are a number of cases where an intra-abstraction policy has to be queried.

\begin{figure}[t]
\centering
\includegraphics[width=1.0\linewidth]{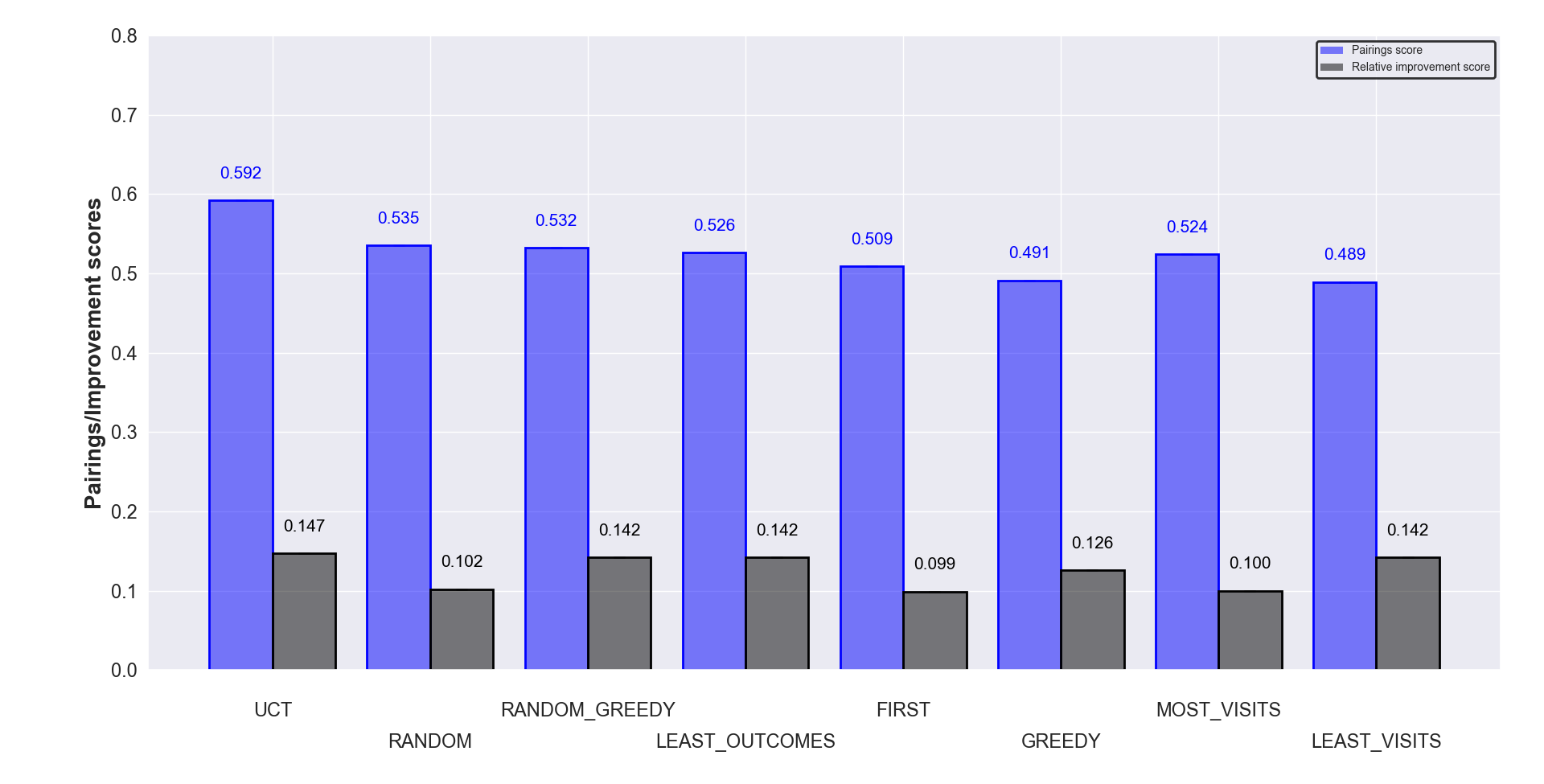}

\caption{The pairings and relative improvement score for all iteration budgets combined of the best performing parameter-combination of each intra-abstraction policy are shown.}
\label{fig:intra:allbudgets}
\end{figure}

\noindent \textbf{Comparison of all intra-abstraction policies:}
Next, we tested which of the intra-abstraction policies performs best overall by determining which policy the best-performing parameter combination uses. In particular, we ran experiments with all intra-abstraction policies on pruned OGA, $(\varepsilon_{\text{a}},\varepsilon_{\text{t}})$-OGA, and RANDOM-OGA (OGA with random abstractions, see \ref{sec:random_oga}, to test the behavior of the intra-abstraction policies on yet another abstraction type). For each abstraction algorithm, we varied $\text{PG} \in \{\text{Yes,No}\}$. For $(\varepsilon_{\text{a}},\varepsilon_{\text{t}})$-OGA, we tested $\varepsilon_{\text{a}} \in \{0,\infty\}$, $\varepsilon_{\text{t}} \in \{0,0.2,0.4,0.8,1.2,1.6\}$, for pruned OGA we used $\alpha \in \{0,0.1,0.2,0.5,0.75,1.0\}$, and for RANDOM-OGA we used $p_{\text{abs}} \in \{0.1,0.2,0.5,1.0\}$. Each parameter combination was run with 100, 200, 500, and 1000 iterations using $C=2$. The bar chart in Fig.~\ref{fig:intra:allbudgets} shows the pairings and relative improvement scores for the parameter combination with the highest scores for each intra-abstraction policy for all iteration budgets combined. The scores for each individual iteration budget are visualized in the supplementary materials in Fig.~\ref{fig:intra:gen}. 
The following key observations can be made.

\noindent 1) First and foremost, the RANDOM intra abstraction policy that has thus far been implicitly used in the literature is always decisively beaten by at least one other strategy. Additionally, the FIRST strategy, which one might accidentally use when no proper tiebreaking is implemented, is even worse. This shows that choosing a suitable intra-abstraction policy is an important aspect when designing an abstraction algorithm.

\noindent 2) Secondly, it is UCT that consistently performs either best or second best in both scores across all budgets, while all other methods fluctuate in their performance, which might be caused due to the few tasks they are calculated over (just a little over 10 environments). The overall best performing parameter-combination was $(0,\infty)$-OGA using $\text{PG}=\text{no}$ and UCT as the intra-abstraction policy (this achieved the values 0.592 and 0.147 in the all-budgets bar chart). The overall best RANDOM-using strategy in terms of the rel. improvement score was pruned OGA with $\alpha=0.75, \text{PG}=\text{no}$ and pruned OGA with $\alpha=0.5, \text{PG}=\text{no}$ for the pairings score (these correspond to the values 0.535 and 0.102\ in the all-budgets bar chart). 

\noindent 3) For the remaining strategies, it is difficult to identify trends that are certainly beyond noise. However, given that RANDOM\_GREEDY only slightly improves over RANDOM in the pairings score shows that, though the intra-abstraction policy at decision time has a considerable performance impact, the tree-policy intra-abstraction policy is equally as important as shown by the performance improvement of UCT.

In sum, the choice of an intra-abstraction policy can have a great impact on performance, which we further consolidate in the next subsection. While the best intra-abstraction policy depends on the concrete iteration budget - model setting, a consistent gain over RANDOM can be gained by simply replacing it with the UCT strategy. In the supplementary materials Section \ref{subsec:intra:ablation}, we show that this drop-in replacement improvement also holds when the abstractions themselves are fixed and one has to find a strategy that can best deal with these abstractions.

\begin{figure}[H]
\centering

\begin{minipage}{0.3\textwidth}
\centering
\includegraphics[width=\linewidth]{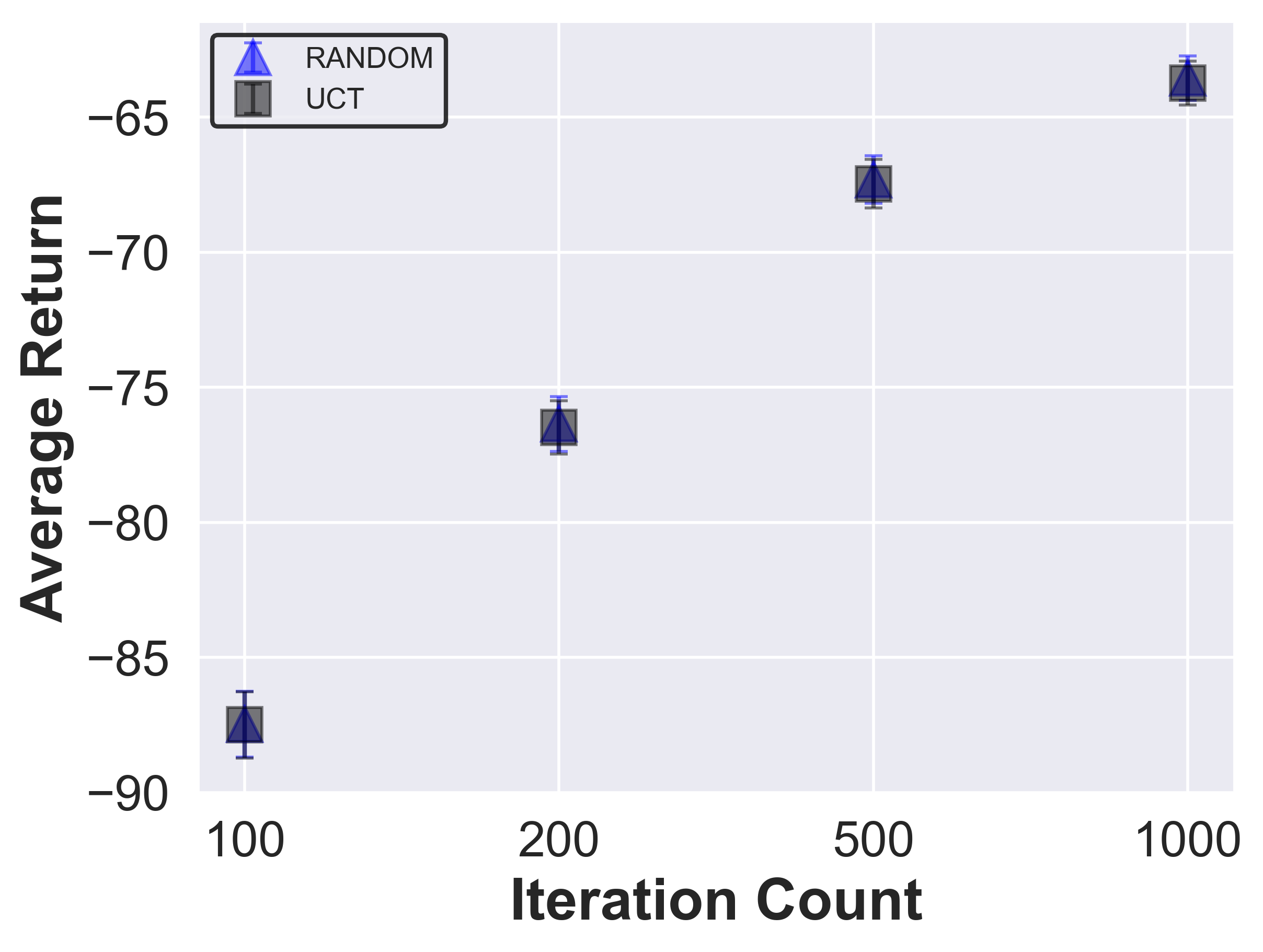}
\caption*{(a) Academic Advising}
\end{minipage}
\hfill
\begin{minipage}{0.3\textwidth}
\centering
\includegraphics[width=\linewidth]{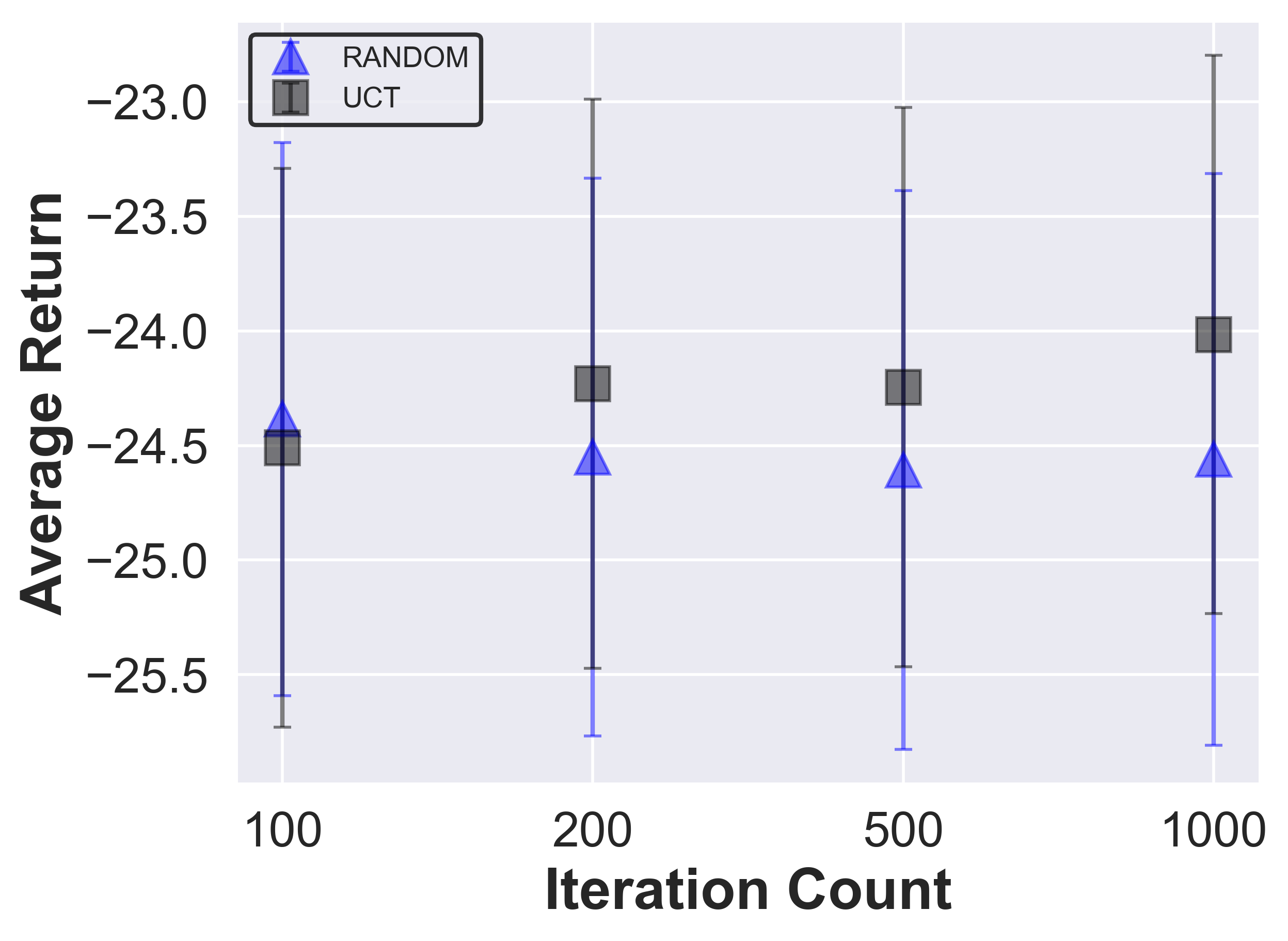}
\caption*{(b) Crossing Traffic}
\end{minipage}
\hfill
\begin{minipage}{0.3\textwidth}
\centering
\includegraphics[width=\linewidth]{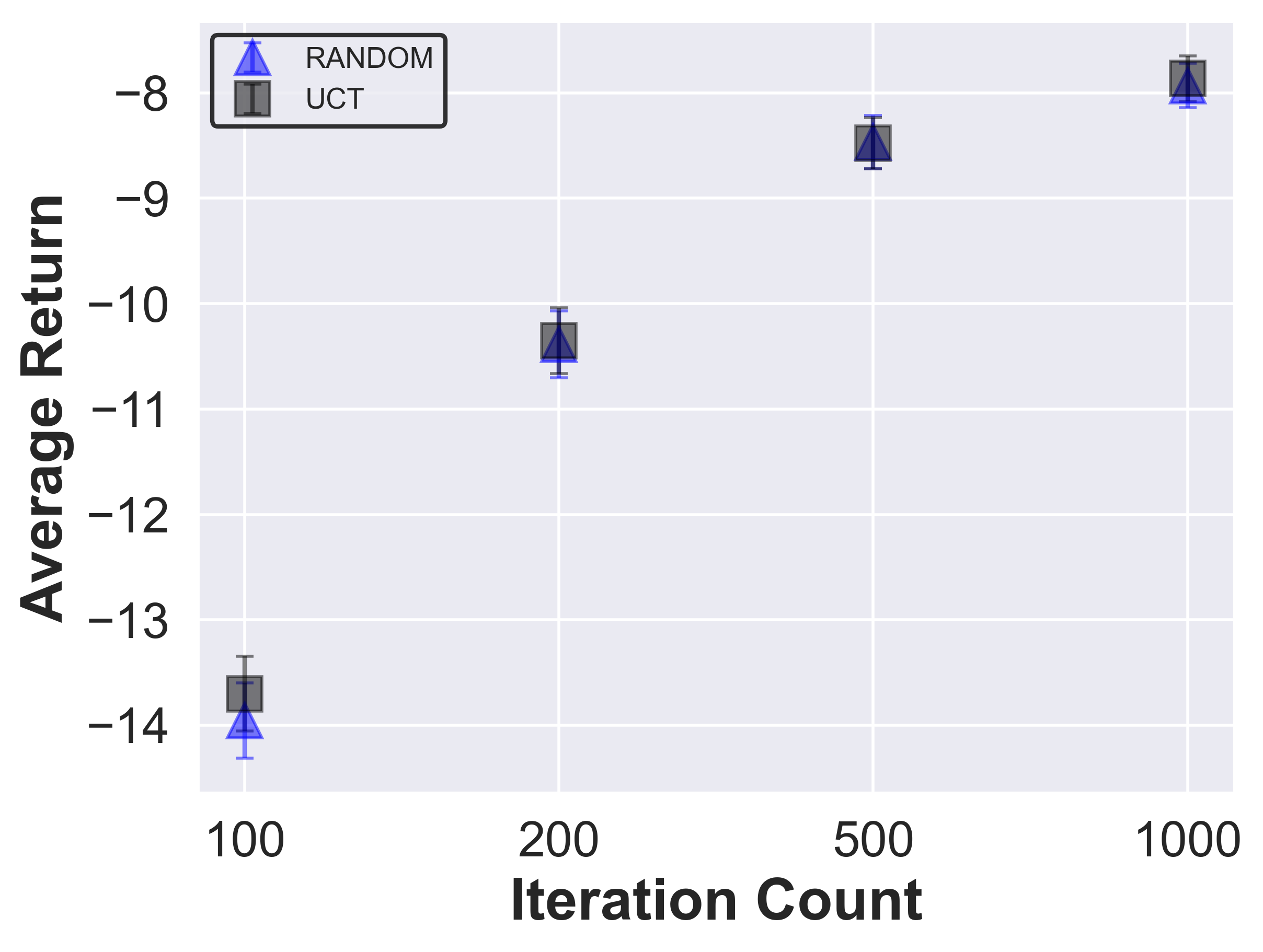}
\caption*{(c) Earth Observation}
\end{minipage}
\hfill
\begin{minipage}{0.3\textwidth}
\centering
\includegraphics[width=\linewidth]{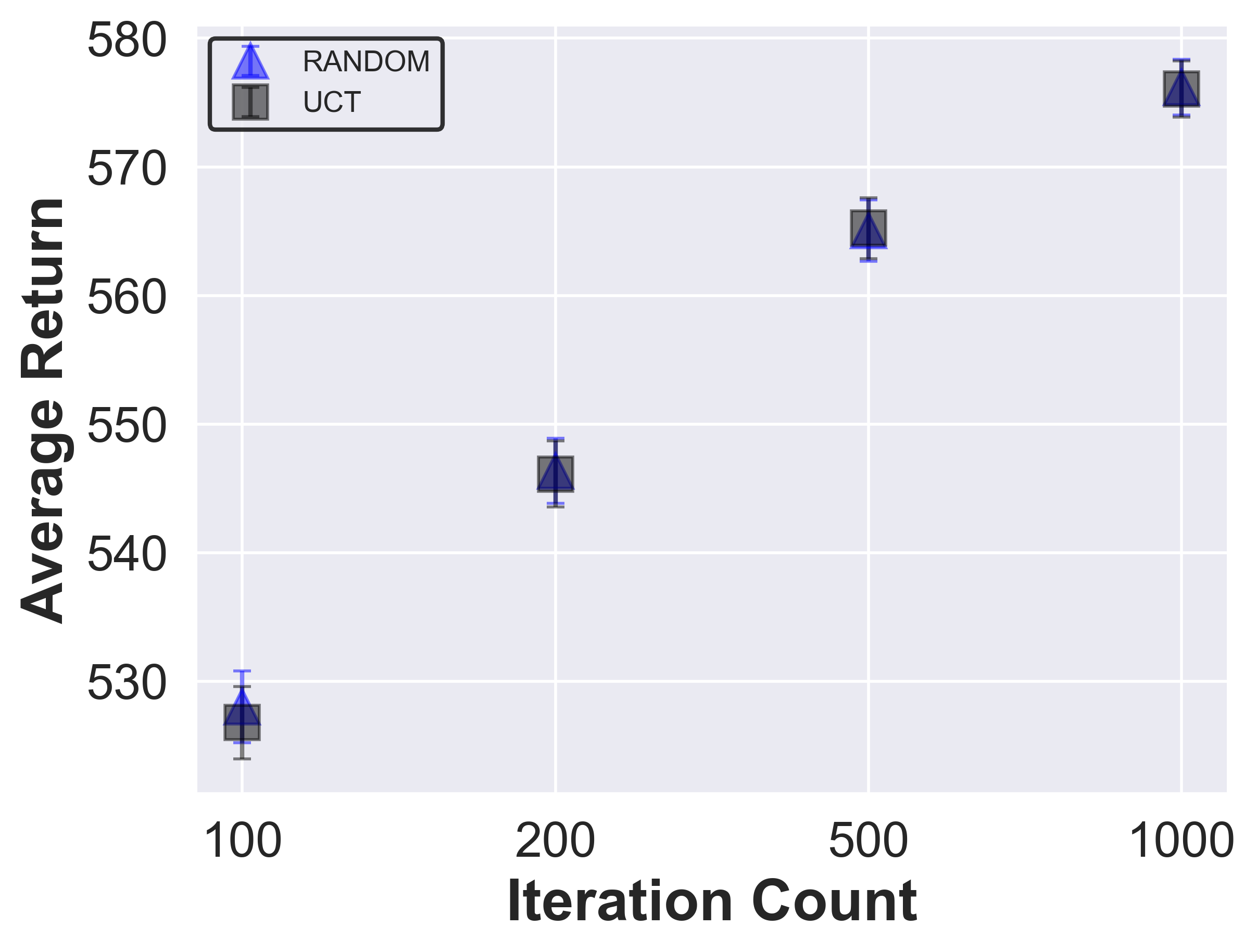}
\caption*{(d) Game of Life}
\end{minipage}
\hfill
\begin{minipage}{0.3\textwidth}
\centering
\includegraphics[width=\linewidth]{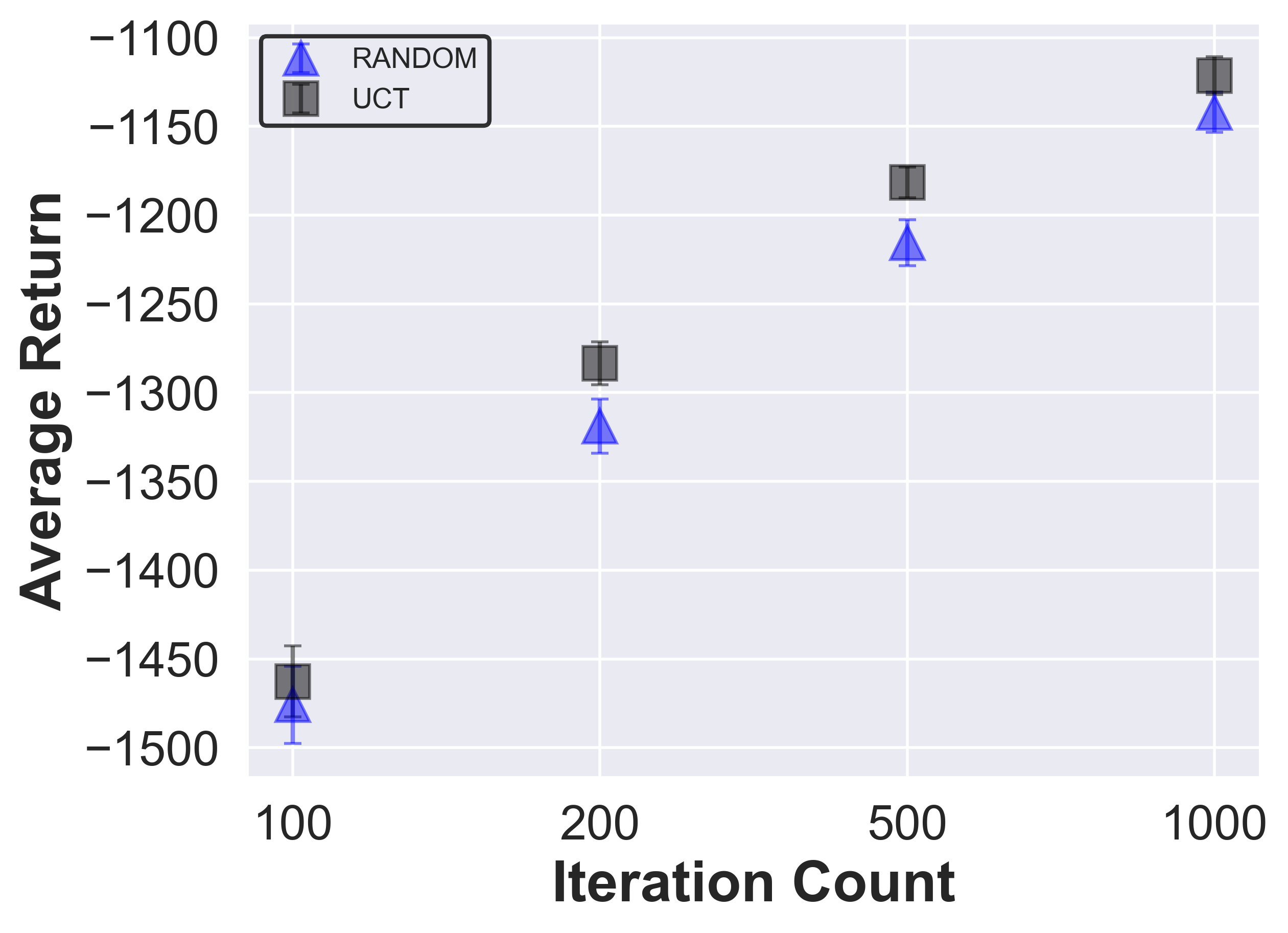}
\caption*{(e) Manufacturer}
\end{minipage}
\hfill
\begin{minipage}{0.3\textwidth}
\centering
\includegraphics[width=\linewidth]{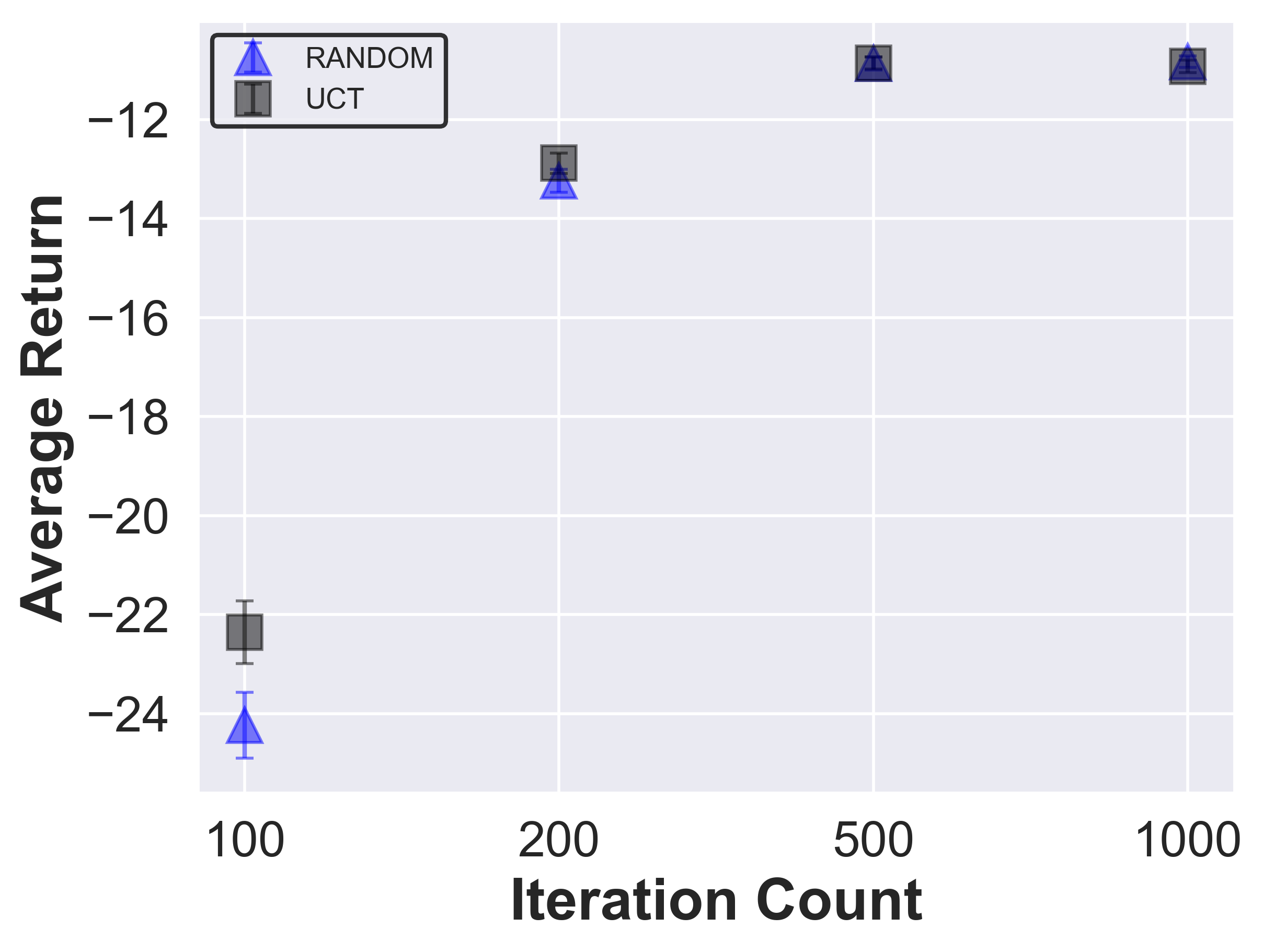}
\caption*{(f) Navigation}
\end{minipage}
\hfill
\begin{minipage}{0.3\textwidth}
\centering
\includegraphics[width=\linewidth]{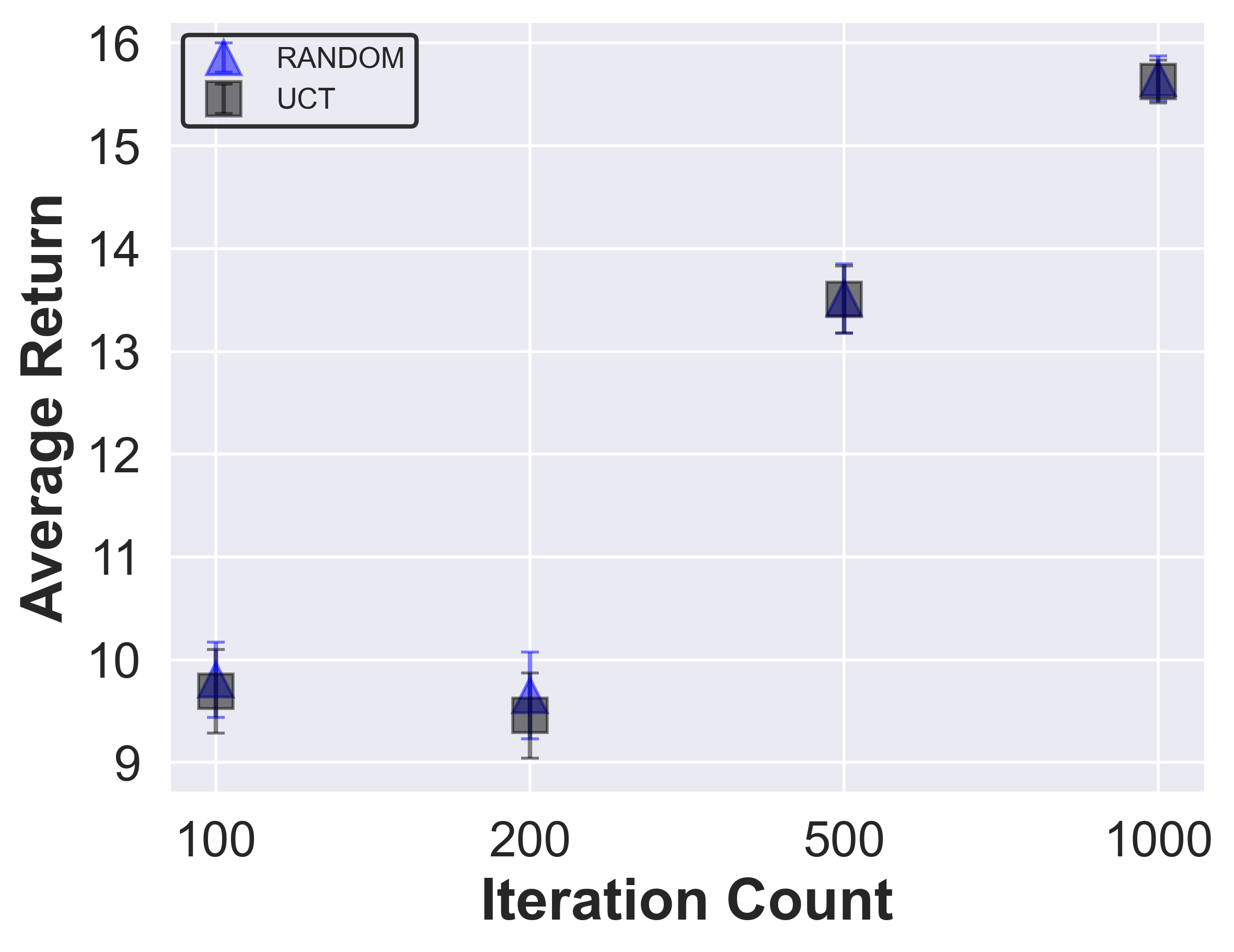}
\caption*{(g) Cooperative Recon}
\end{minipage}
\hfill
\begin{minipage}{0.3\textwidth}
\centering
\includegraphics[width=\linewidth]{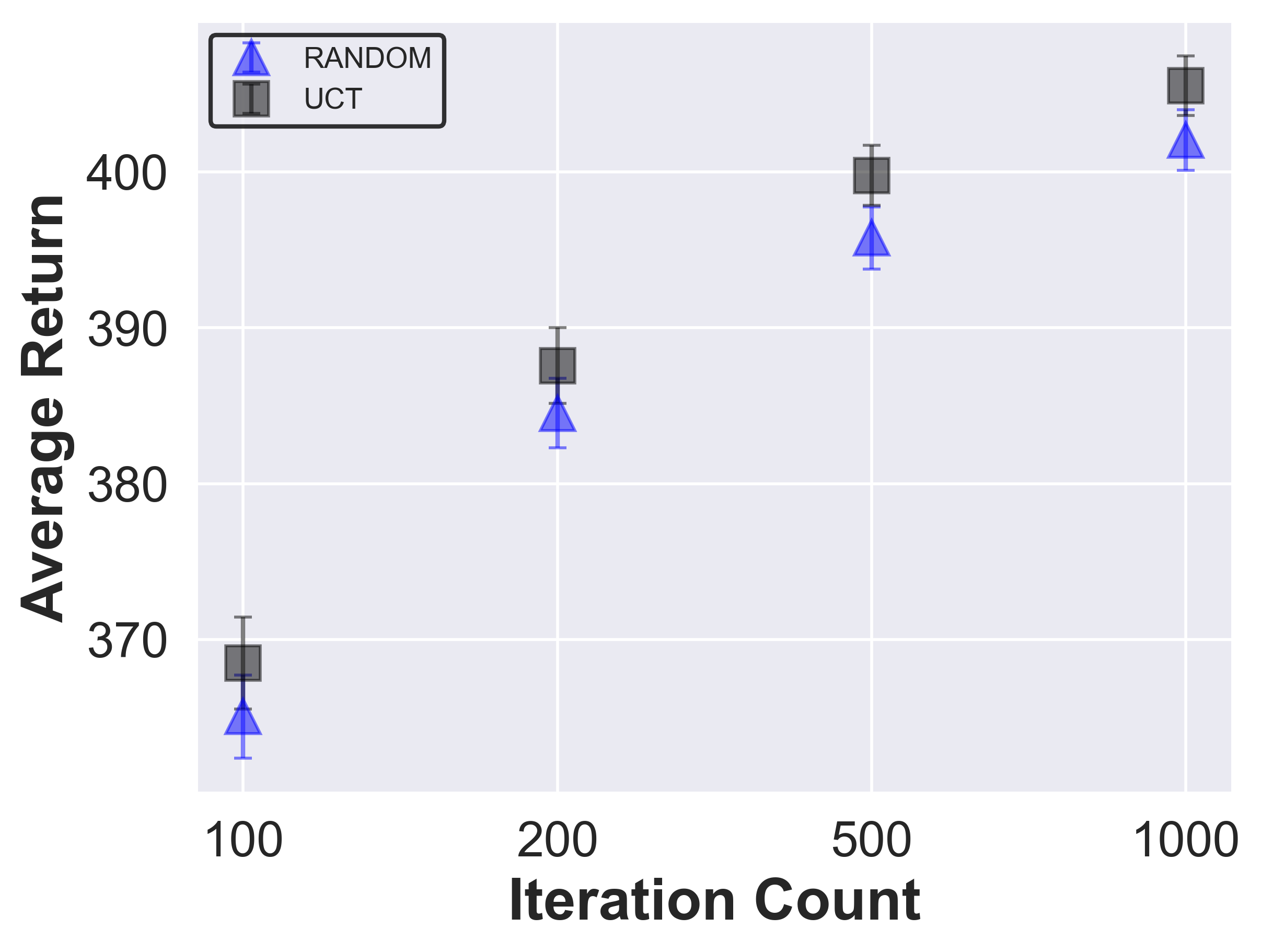}
\caption*{(h) SysAdmin}
\end{minipage}
\hfill
\begin{minipage}{0.3\textwidth}
\centering
\includegraphics[width=\linewidth]{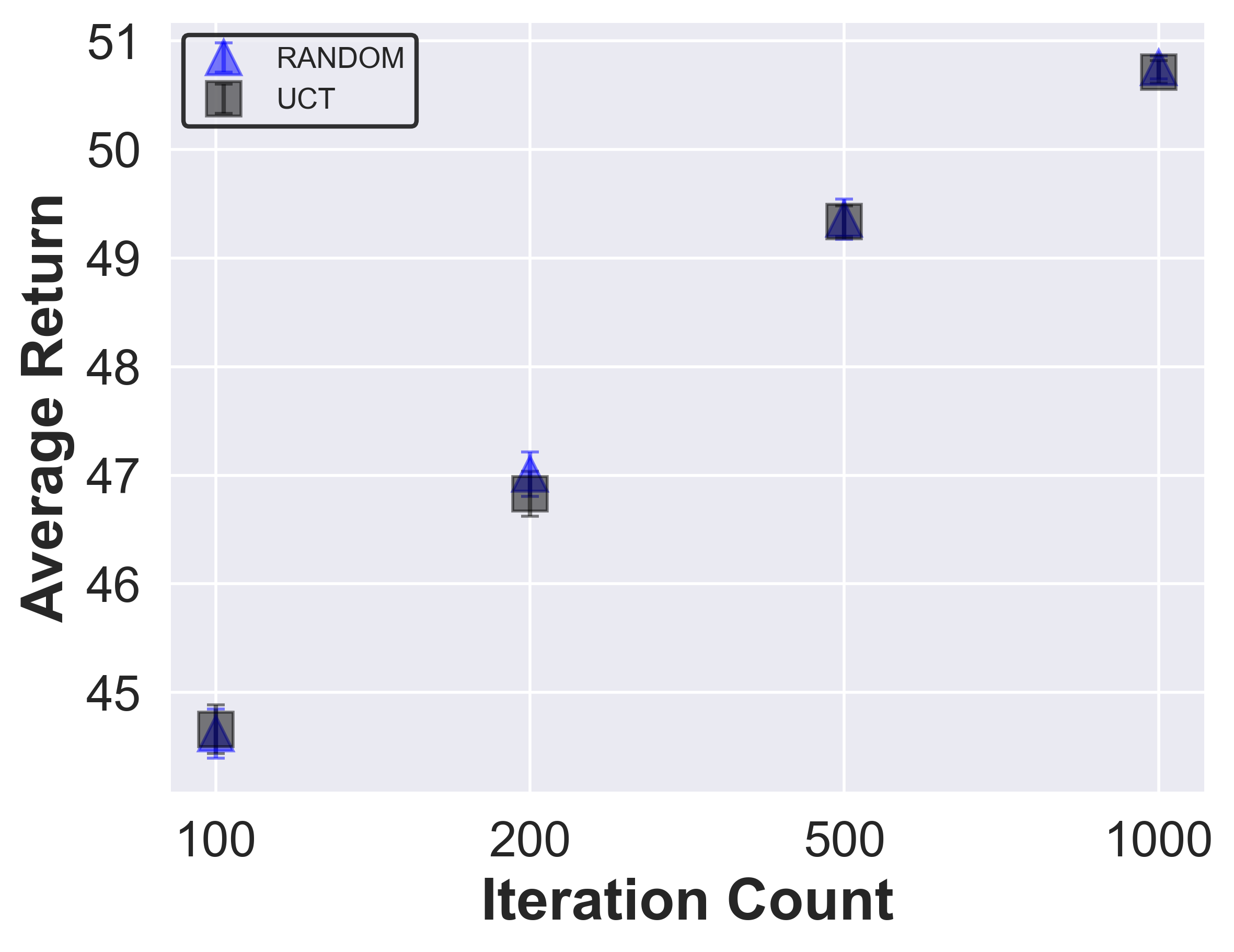}
\caption*{(i) Saving}
\end{minipage}
\hfill
\begin{minipage}{0.3\textwidth}
\centering
\includegraphics[width=\linewidth]{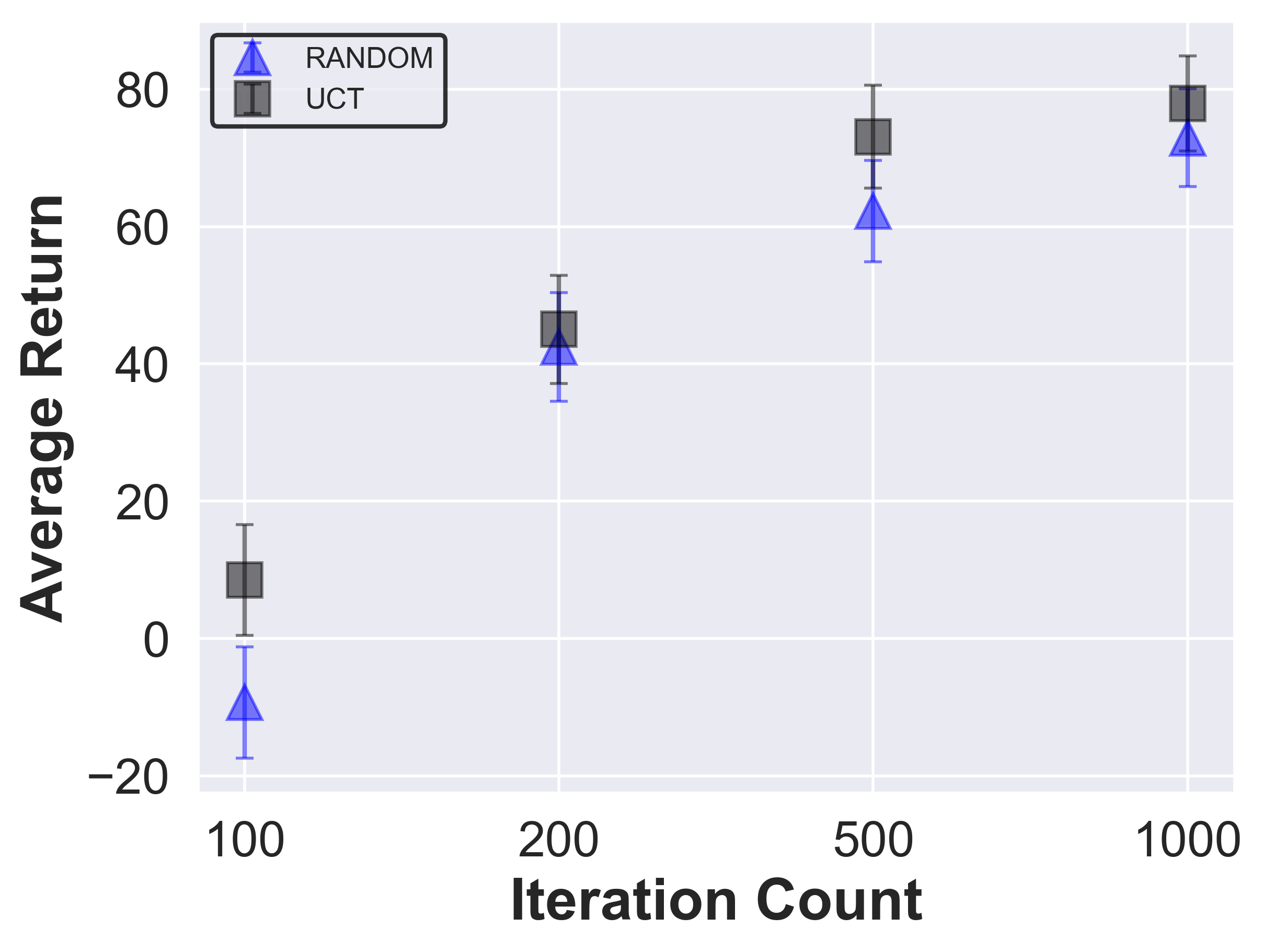}
\caption*{(j) Skill Teaching}
\end{minipage}
\hfill
\begin{minipage}{0.3\textwidth}
\centering
\includegraphics[width=\linewidth]{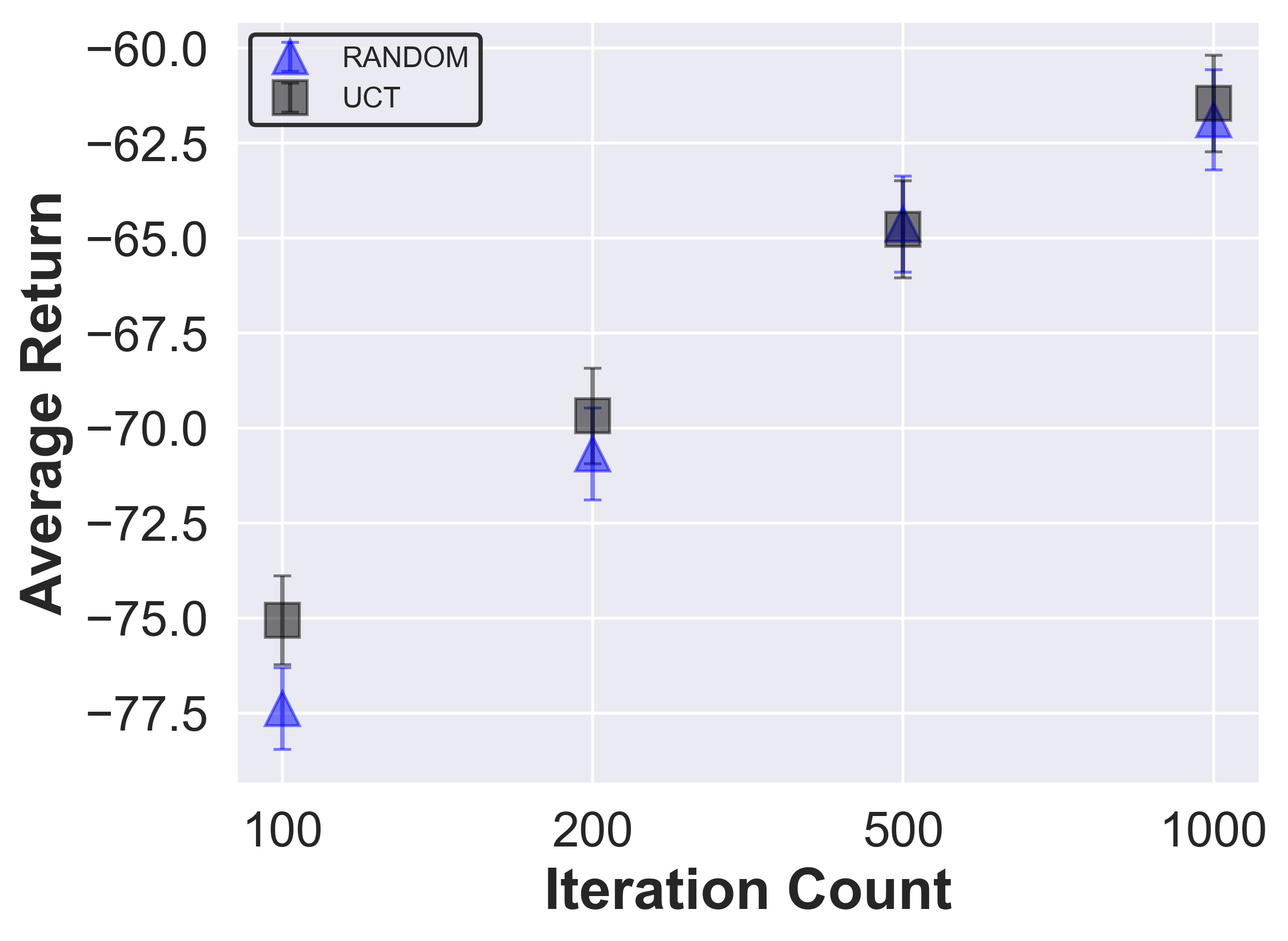}
\caption*{(k) Sailing Wind}
\end{minipage}
\hfill
\begin{minipage}{0.3\textwidth}
\centering
\includegraphics[width=\linewidth]{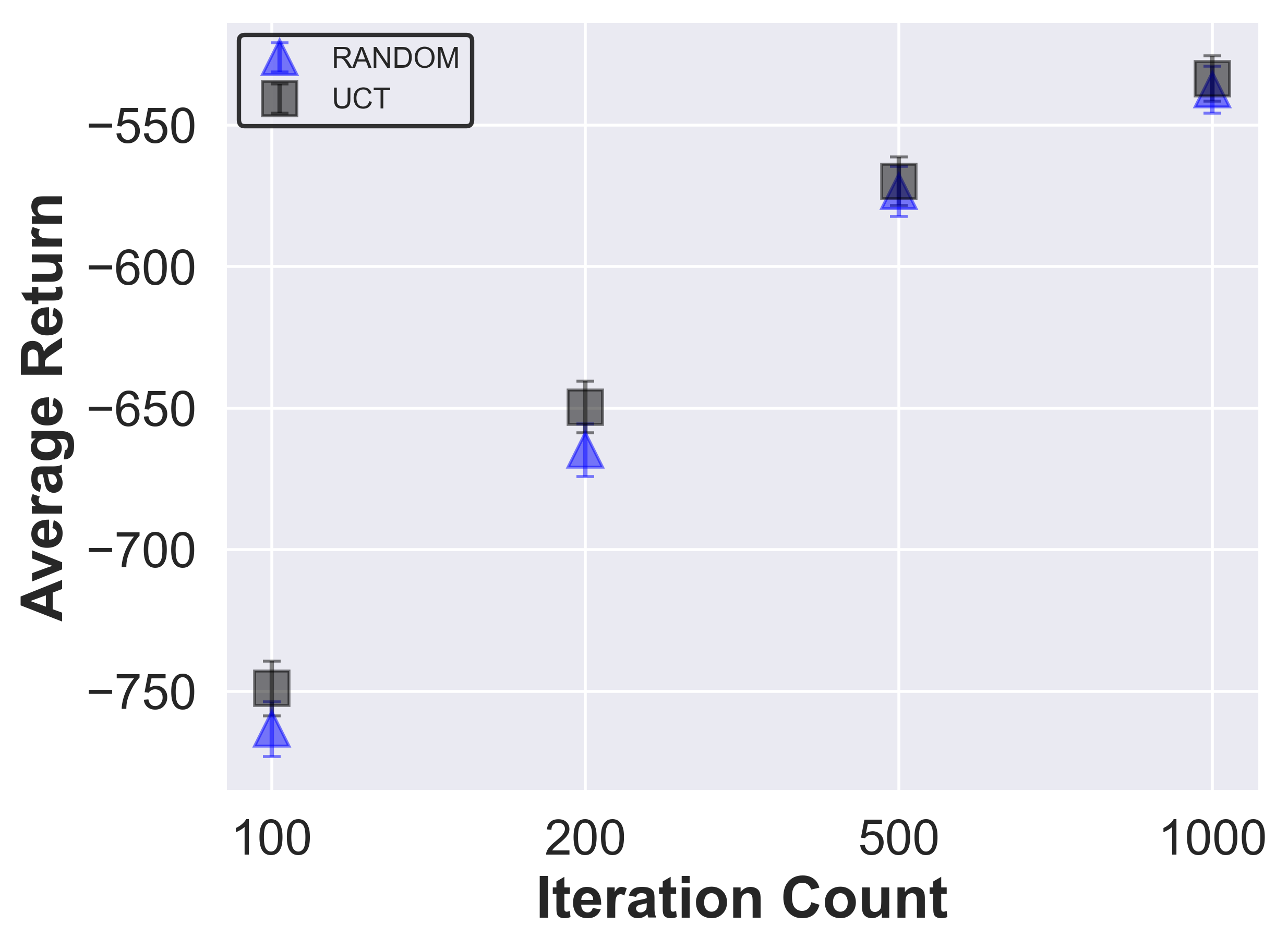}
\caption*{(l) Tamarisk}
\end{minipage}
\hfill
\begin{minipage}{0.3\textwidth}
\centering
\includegraphics[width=\linewidth]{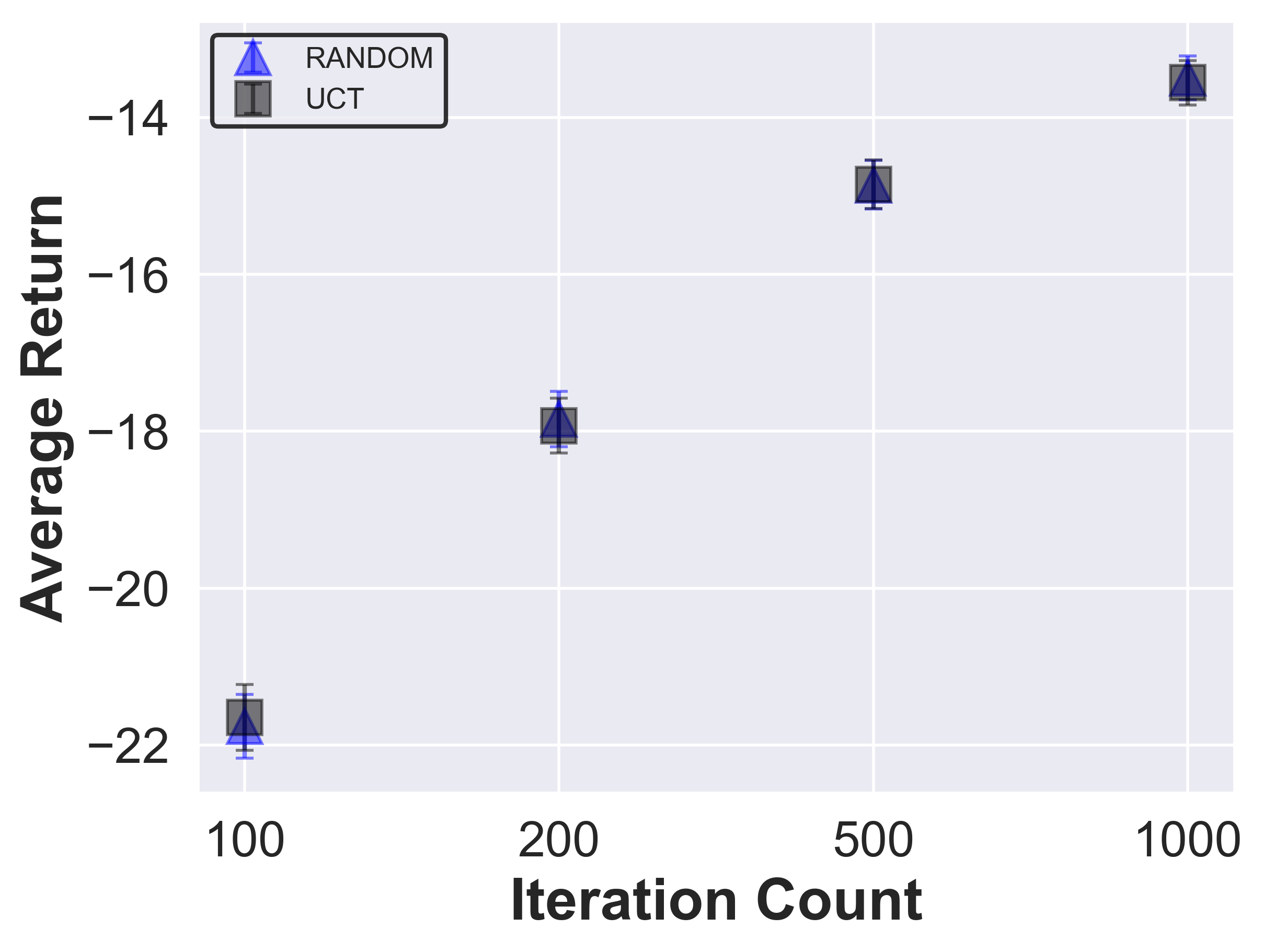}
\caption*{(m) Traffic}
\end{minipage}
\hfill
\begin{minipage}{0.3\textwidth}
\centering
\includegraphics[width=\linewidth]{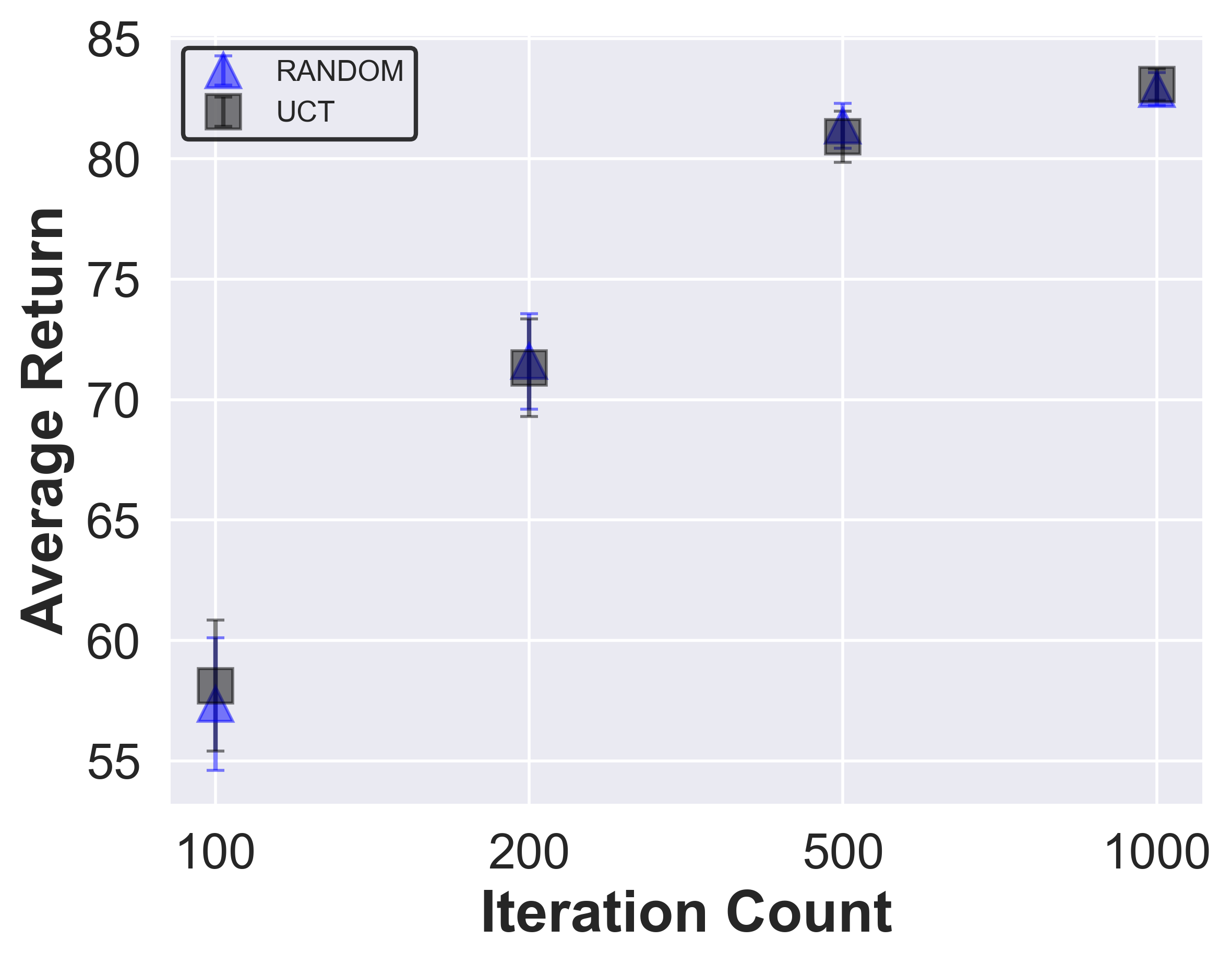}
\caption*{(m) Triangle Tireworld}
\end{minipage}
\hfill
\begin{minipage}{0.3\textwidth}
\centering
\includegraphics[width=\linewidth]{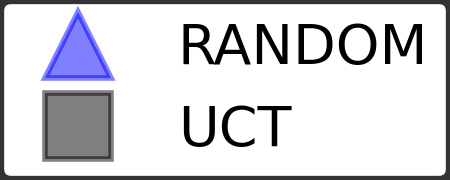}
\caption*{Legend}
\end{minipage}

\caption{The performance graphs of in dependence of the MCTS iteration count of the parameter optimized versions of pruned OGA, $(\varepsilon_{\text{a}},\varepsilon_{\text{t}})$-OGA, and RANDOM-OGA combined with the UCT or RANDOM intra-abstraction policy.}
\label{fig:intra:optimized}
\end{figure}

\noindent \textbf{Parameter-optimized performances:}
Next, we compared RANDOM versus the UCT intra abstraction policy in the parameter-optimized setting, where we optimized both agents over the same set of parameters in addition to varying the exploration constant in $C \in \{0.5,1,2,4,8,16\}$ and including additional domain-specific $\varepsilon_{\text{a}}$ values that are listed in Tab.~\ref{tab:ogaeps:epsa_values} in the supplementary materials. For feasibility reasons, we restricted ourselves to UCT instead of additionally including all other intra-abstraction policies.
Fig.~\ref{fig:intra:optimized} shows the performance graphs for each environment. The following observations can be made.

\noindent 1) In every environment UCT either clearly performs better than RANDOM with at least one iteration budget (in Crossing Traffic, Earth Observation, Manufacturer, Navigation, SysAdmin, Skill Teaching, Sailing Wind and Tamarisk) or performs on par (in Academic Advising, Game of Life, Cooperative Recon, Saving, Traffic, and Triangle Tireworld). 

\noindent 2) The performance gains can be explained by the fact that UCT performs better (relative to RANDOM), the coarser the abstraction, as shown in the supplementary materials Section \ref{subsec:intra:coarsenesses}. The environments where performance is gained over RANDOM are those that satisfy the following criteria. Firstly, using coarse abstractions is either on par with OGA-UCT or even better. Secondly, the coarseness needs to introduce actual abstraction errors which for example, is rarely the case in Game of Life. Lastly, the intra-abstraction policy needs to be queried in the first place, which explains why there is no gain for Navigation at higher iteration budgets. 

In summary, using intra-abstraction policies is a valuable tool to improve peak performances across a wide range of environments, especially when the peak performance without intra-abstraction policies has been reached with a coarse abstraction.

\section{Limitations and Future Work}
\label{sec:future_work}
In this paper, we first generalized the ASAP and AS frameworks to ASASAP. We then brought attention to the intra-abstraction policy problem and showed that this is not an edge case. To relieve this issue, we proposed several intra-abstraction policies as an alternative to the random policy that is implicitly used in standard OGA. While some of them were only marginally better than RANDOM, like MIN\_VISITS and RANDOM\_GREEDY, we found that UCT-OGA performs best and consistently, performing either on par or clearly outperforming standard OGA across a variety of parameter settings and environments. Consequently, we believe that UCT should be used as the standard intra-abstraction policy for MCTS-based abstraction methods instead of the random policy.

Firstly, limitations of OGA also directly translate to OGA enhanced with an intra-abstraction policy. In particular, for any performance gains to appear in the first place, the environment must contain state-action pairs with the same Q value. Furthermore, for any abstractions to be detected in the first place, the search graph must be a directed acyclic graph to ensure that there are state-action pairs with the same successors, a necessary condition for any abstractions (unless $\varepsilon_{\text{t}}=2$).
Another intra-abstraction policy-specific limitation is that these are only useful for non-exact abstractions. In particular, if only state-action pairs with the same optimal Q value are abstracted then which one the intra-abstraction policy chooses makes no difference.

From a different viewpoint, intra-abstraction policies can be seen as some special form of operating in hierarchical abstractions, where one iteratively selects actions from different layers of abstractions until a ground action is reached. In the case of intra-abstraction policies, the hierarchy consists of two abstractions: the standard ASAP abstraction, followed by a trivial one where every node is assigned to its own abstract node. We believe that choosing an intra-abstraction policy that itself just selects an abstract action from a finer abstraction could be worth investigating.

Whenever the intra-abstraction policies resulted in ties, these were always resolved randomly. Even though any gains here would be even more marginal, one could probably find further optimizations by setting up a tiebreak hierarchy, e.g., if UCT results in a tie, then these are resolved by MOST\_VISITS.

Furthermore, on a more general level, it might be worth investigating if the progress in search abstractions can be translated to machine learning methods that are built on these searches, such as AlphaZero \citep{alphazero}.

\newpage 

\bibliography{references}

\begin{thebibliography}{29}
\providecommand{\natexlab}[1]{#1}
\providecommand{\url}[1]{\texttt{#1}}
\expandafter\ifx\csname urlstyle\endcsname\relax
  \providecommand{\doi}[1]{doi: #1}\else
  \providecommand{\doi}{doi: \begingroup \urlstyle{rm}\Url}\fi

\bibitem[Anand et~al.(2015)Anand, Grover, Mausam, and Singla]{AnandGMS15}
Ankit Anand, Aditya Grover, Mausam, and Parag Singla.
\newblock {{ASAP-UCT:} Abstraction of State-Action Pairs in {UCT}}.
\newblock In Qiang Yang and Michael~J. Wooldridge (eds.), \emph{Proceedings of the Twenty-Fourth International Joint Conference on Artificial Intelligence, {IJCAI} 2015, Buenos Aires, Argentina, July 25-31, 2015}, pp.\  1509--1515. {AAAI} Press, 2015.
\newblock URL \url{http://ijcai.org/Abstract/15/216}.

\bibitem[Anand et~al.(2016)Anand, Noothigattu, Mausam, and Singla]{OGAUCT}
Ankit Anand, Ritesh Noothigattu, Mausam, and Parag Singla.
\newblock {OGA-UCT: on-the-go abstractions in UCT}.
\newblock In \emph{Proceedings of the Twenty-Sixth International Conference on International Conference on Automated Planning and Scheduling}, ICAPS'16, pp.\  29–37. AAAI Press, 2016.
\newblock ISBN 1577357574.

\bibitem[Berner et~al.(2019)Berner, Brockman, Chan, Cheung, Debiak, Dennison, Farhi, Fischer, Hashme, Hesse, J{\'{o}}zefowicz, Gray, Olsson, Pachocki, Petrov, de~Oliveira~Pinto, Raiman, Salimans, Schlatter, Schneider, Sidor, Sutskever, Tang, Wolski, and Zhang]{dota2openaifive}
Christopher Berner, Greg Brockman, Brooke Chan, Vicki Cheung, Przemyslaw Debiak, Christy Dennison, David Farhi, Quirin Fischer, Shariq Hashme, Christopher Hesse, Rafal J{\'{o}}zefowicz, Scott Gray, Catherine Olsson, Jakub Pachocki, Michael Petrov, Henrique~Pond{\'{e}} de~Oliveira~Pinto, Jonathan Raiman, Tim Salimans, Jeremy Schlatter, Jonas Schneider, Szymon Sidor, Ilya Sutskever, Jie Tang, Filip Wolski, and Susan Zhang.
\newblock {Dota 2 with Large Scale Deep Reinforcement Learning}.
\newblock \emph{CoRR}, abs/1912.06680, 2019.
\newblock URL \url{http://arxiv.org/abs/1912.06680}.

\bibitem[Browne et~al.(2012)Browne, Powley, Whitehouse, Lucas, Cowling, Rohlfshagen, Tavener, Liebana, Samothrakis, and Colton]{BrownePWLCRTPSC12}
Cameron Browne, Edward~Jack Powley, Daniel Whitehouse, Simon~M. Lucas, Peter~I. Cowling, Philipp Rohlfshagen, Stephen Tavener, Diego~Perez Liebana, Spyridon Samothrakis, and Simon Colton.
\newblock {A Survey of Monte Carlo Tree Search Methods}.
\newblock \emph{{IEEE} Trans. Comput. Intell. {AI} Games}, 4\penalty0 (1):\penalty0 1--43, 2012.
\newblock \doi{10.1109/TCIAIG.2012.2186810}.
\newblock URL \url{https://doi.org/10.1109/TCIAIG.2012.2186810}.

\bibitem[Chitnis et~al.(2020)Chitnis, Silver, Kim, Kaelbling, and Lozano{-}P{\'{e}}rez]{ChitnisSKKL20}
Rohan Chitnis, Tom Silver, Beomjoon Kim, Leslie~Pack Kaelbling, and Tom{\'{a}}s Lozano{-}P{\'{e}}rez.
\newblock Camps: Learning context-specific abstractions for efficient planning in factored mdps.
\newblock In Jens Kober, Fabio Ramos, and Claire~J. Tomlin (eds.), \emph{4th Conference on Robot Learning, CoRL 2020, 16-18 November 2020, Virtual Event / Cambridge, MA, {USA}}, volume 155 of \emph{Proceedings of Machine Learning Research}, pp.\  64--79. {PMLR}, 2020.
\newblock URL \url{https://proceedings.mlr.press/v155/chitnis21a.html}.

\bibitem[Coulom(2006)]{coulom06}
R{\'{e}}mi Coulom.
\newblock Efficient selectivity and backup operators in monte-carlo tree search.
\newblock In H.~Jaap van~den Herik, Paolo Ciancarini, and H.~H. L.~M. Donkers (eds.), \emph{Computers and Games, 5th International Conference, {CG} 2006, Turin, Italy, May 29-31, 2006. Revised Papers}, volume 4630 of \emph{Lecture Notes in Computer Science}, pp.\  72--83. Springer, 2006.
\newblock \doi{10.1007/978-3-540-75538-8\_7}.
\newblock URL \url{https://doi.org/10.1007/978-3-540-75538-8\_7}.

\bibitem[Eyck \& M{\"{u}}ller(2011)Eyck and M{\"{u}}ller]{EyckM11}
Gabriel~Van Eyck and Martin M{\"{u}}ller.
\newblock {Revisiting Move Groups in Monte-Carlo Tree Search}.
\newblock In H.~Jaap van~den Herik and Aske Plaat (eds.), \emph{Advances in Computer Games - 13th International Conference, {ACG} 2011, Tilburg, The Netherlands, November 20-22, 2011, Revised Selected Papers}, volume 7168 of \emph{Lecture Notes in Computer Science}, pp.\  13--23. Springer, 2011.
\newblock \doi{10.1007/978-3-642-31866-5\_2}.
\newblock URL \url{https://doi.org/10.1007/978-3-642-31866-5\_2}.

\bibitem[Givan et~al.(2003)Givan, Dean, and Greig]{GivanDG03}
Robert Givan, Thomas~L. Dean, and Matthew Greig.
\newblock {Equivalence notions and model minimization in Markov decision processes}.
\newblock \emph{Artif. Intell.}, 147\penalty0 (1-2):\penalty0 163--223, 2003.
\newblock \doi{10.1016/S0004-3702(02)00376-4}.
\newblock URL \url{https://doi.org/10.1016/S0004-3702(02)00376-4}.

\bibitem[Grzes et~al.(2014)Grzes, Hoey, and Sanner]{grzes2014ippc}
Marek Grzes, Jesse Hoey, and Scott Sanner.
\newblock {{International Probabilistic Planning Competition (IPPC) 2014}}.
\newblock In \emph{Proceedings of the International Conference on Automated Planning and Scheduling (ICAPS)}, 2014.

\bibitem[Hoerger et~al.(2024)Hoerger, Kurniawati, Kroese, and Ye]{HoergerKKY24}
Marcus Hoerger, Hanna Kurniawati, Dirk~P. Kroese, and Nan Ye.
\newblock Adaptive discretization using voronoi trees for continuous pomdps.
\newblock \emph{Int. J. Robotics Res.}, 43\penalty0 (9):\penalty0 1283--1298, 2024.
\newblock \doi{10.1177/02783649231188984}.
\newblock URL \url{https://doi.org/10.1177/02783649231188984}.

\bibitem[Hostetler et~al.(2015)Hostetler, Fern, and Dietterich]{HostetlerFD15}
Jesse Hostetler, Alan Fern, and Thomas~G. Dietterich.
\newblock {Progressive Abstraction Refinement for Sparse Sampling}.
\newblock In Marina Meila and Tom Heskes (eds.), \emph{Proceedings of the Thirty-First Conference on Uncertainty in Artificial Intelligence, {UAI} 2015, July 12-16, 2015, Amsterdam, The Netherlands}, pp.\  365--374. {AUAI} Press, 2015.
\newblock URL \url{http://auai.org/uai2015/proceedings/papers/81.pdf}.

\bibitem[Jiang et~al.(2014)Jiang, Singh, and Lewis]{uctJiang}
Nan Jiang, Satinder Singh, and Richard~L. Lewis.
\newblock {Improving {UCT} planning via approximate homomorphisms}.
\newblock In Ana L.~C. Bazzan, Michael~N. Huhns, Alessio Lomuscio, and Paul Scerri (eds.), \emph{International conference on Autonomous Agents and Multi-Agent Systems, {AAMAS} '14, Paris, France, May 5-9, 2014}, pp.\  1289--1296. {IFAAMAS/ACM}, 2014.
\newblock URL \url{http://dl.acm.org/citation.cfm?id=2617453}.

\bibitem[Kocsis \& Szepesv{\'{a}}ri(2006)Kocsis and Szepesv{\'{a}}ri]{KocsisS06}
Levente Kocsis and Csaba Szepesv{\'{a}}ri.
\newblock {Bandit Based Monte-Carlo Planning}.
\newblock In Johannes F{\"{u}}rnkranz, Tobias Scheffer, and Myra Spiliopoulou (eds.), \emph{Machine Learning: {ECML} 2006, 17th European Conference on Machine Learning, Berlin, Germany, September 18-22, 2006, Proceedings}, volume 4212 of \emph{Lecture Notes in Computer Science}, pp.\  282--293. Springer, 2006.
\newblock \doi{10.1007/11871842\_29}.
\newblock URL \url{https://doi.org/10.1007/11871842\_29}.

\bibitem[Kwak et~al.(2024)Kwak, Hwang, Kim, Lee, and Zhang]{KwakHKLZ24}
Yunhyeok Kwak, Inwoo Hwang, Dooyoung Kim, Sanghack Lee, and Byoung{-}Tak Zhang.
\newblock Efficient monte carlo tree search via on-the-fly state-conditioned action abstraction.
\newblock In Negar Kiyavash and Joris~M. Mooij (eds.), \emph{Uncertainty in Artificial Intelligence, 15-19 July 2024, Universitat Pompeu Fabra, Barcelona, Spain}, volume 244 of \emph{Proceedings of Machine Learning Research}, pp.\  2076--2093. {PMLR}, 2024.
\newblock URL \url{https://proceedings.mlr.press/v244/kwak24a.html}.

\bibitem[Ozair et~al.(2021)Ozair, Li, Razavi, Antonoglou, van~den Oord, and Vinyals]{OzairLRAOV21}
Sherjil Ozair, Yazhe Li, Ali Razavi, Ioannis Antonoglou, A{\"{a}}ron van~den Oord, and Oriol Vinyals.
\newblock Vector quantized models for planning.
\newblock In Marina Meila and Tong Zhang (eds.), \emph{Proceedings of the 38th International Conference on Machine Learning, {ICML} 2021, 18-24 July 2021, Virtual Event}, volume 139 of \emph{Proceedings of Machine Learning Research}, pp.\  8302--8313. {PMLR}, 2021.
\newblock URL \url{http://proceedings.mlr.press/v139/ozair21a.html}.

\bibitem[Ravindran \& Barto(2004)Ravindran and Barto]{ravindran2004approximate}
B.~Ravindran and A.~G. Barto.
\newblock {Approximate Homomorphisms: A Framework for Non-Exact Minimization in Markov Decision Processes}.
\newblock In \emph{Proc. Int. Conf. Knowl.-Based Comput. Syst.}, pp.\  1--10, 2004.

\bibitem[Saisubramanian et~al.(2017)Saisubramanian, Zilberstein, and Shenoy]{saisubramanian2017optimizing}
S.~Saisubramanian, S.~Zilberstein, and P.~Shenoy.
\newblock {Optimizing Electric Vehicle Charging Through Determinization}.
\newblock In \emph{ICAPS Workshop on Scheduling and Planning Applications}, 2017.

\bibitem[Schmöcker(2025)]{repo}
Robin Schmöcker.
\newblock {IntraAbsPolicies}, 2025.
\newblock Repository available at: \url{https://github.com/codebro634/IntraAbsPolicies.git}.

\bibitem[Schmöcker \& Dockhorn(2025)Schmöcker and Dockhorn]{mysurvey}
Robin Schmöcker and Alexander Dockhorn.
\newblock A survey of non-learning-based abstractions for sequential decision-making.
\newblock \emph{IEEE Access}, 13:\penalty0 100808--100830, 2025.
\newblock \doi{10.1109/ACCESS.2025.3572830}.

\bibitem[Schmöcker et~al.(2025{\natexlab{a}})Schmöcker, Dockhorn, and Rosenhahn]{aupo}
Robin Schmöcker, Alexander Dockhorn, and Bodo Rosenhahn.
\newblock Aupo - abstracted until proven otherwise: A reward distribution based abstraction algorithm, 2025{\natexlab{a}}.
\newblock URL \url{https://arxiv.org/abs/2510.23214}.

\bibitem[Schmöcker et~al.(2025{\natexlab{b}})Schmöcker, Kampmann, and Dockhorn]{ogacad}
Robin Schmöcker, Lennart Kampmann, and Alexander Dockhorn.
\newblock Time-critical and confidence-based abstraction dropping methods.
\newblock In \emph{2025 IEEE Conference on Games (CoG)}, pp.\  1--8, 2025{\natexlab{b}}.
\newblock \doi{10.1109/CoG64752.2025.11114261}.

\bibitem[Schmöcker et~al.(2025{\natexlab{c}})Schmöcker, Schnell, and Dockhorn]{demcts}
Robin Schmöcker, Christoph Schnell, and Alexander Dockhorn.
\newblock Investigating scale independent uct exploration factor strategies, 2025{\natexlab{c}}.
\newblock URL \url{https://arxiv.org/abs/2510.21275}.

\bibitem[Silver et~al.(2016)Silver, Huang, Maddison, Guez, Sifre, van~den Driessche, Schrittwieser, Antonoglou, Panneershelvam, Lanctot, Dieleman, Grewe, Nham, Kalchbrenner, Sutskever, Lillicrap, Leach, Kavukcuoglu, Graepel, and Hassabis]{SilverHMGSDSAPL16}
David Silver, Aja Huang, Chris~J. Maddison, Arthur Guez, Laurent Sifre, George van~den Driessche, Julian Schrittwieser, Ioannis Antonoglou, Vedavyas Panneershelvam, Marc Lanctot, Sander Dieleman, Dominik Grewe, John Nham, Nal Kalchbrenner, Ilya Sutskever, Timothy~P. Lillicrap, Madeleine Leach, Koray Kavukcuoglu, Thore Graepel, and Demis Hassabis.
\newblock {Mastering the game of Go with deep neural networks and tree search}.
\newblock \emph{Nat.}, 529\penalty0 (7587):\penalty0 484--489, 2016.
\newblock \doi{10.1038/NATURE16961}.
\newblock URL \url{https://doi.org/10.1038/nature16961}.

\bibitem[Silver et~al.(2017)Silver, Hubert, Schrittwieser, Antonoglou, Lai, Guez, Lanctot, Sifre, Kumaran, Graepel, Lillicrap, Simonyan, and Hassabis]{alphazero}
David Silver, Thomas Hubert, Julian Schrittwieser, Ioannis Antonoglou, Matthew Lai, Arthur Guez, Marc Lanctot, Laurent Sifre, Dharshan Kumaran, Thore Graepel, Timothy~P. Lillicrap, Karen Simonyan, and Demis Hassabis.
\newblock {Mastering Chess and Shogi by Self-Play with a General Reinforcement Learning Algorithm}.
\newblock \emph{CoRR}, abs/1712.01815, 2017.
\newblock URL \url{http://arxiv.org/abs/1712.01815}.

\bibitem[Sokota et~al.(2021)Sokota, Ho, Ahmad, and Kolter]{SokotaHAK21}
Samuel Sokota, Caleb Ho, Zaheen~Farraz Ahmad, and J.~Zico Kolter.
\newblock {Monte Carlo Tree Search With Iteratively Refining State Abstractions}.
\newblock In Marc'Aurelio Ranzato, Alina Beygelzimer, Yann~N. Dauphin, Percy Liang, and Jennifer~Wortman Vaughan (eds.), \emph{Advances in Neural Information Processing Systems 34: Annual Conference on Neural Information Processing Systems 2021, NeurIPS 2021, December 6-14, 2021, virtual}, pp.\  18698--18709, 2021.
\newblock URL \url{https://proceedings.neurips.cc/paper/2021/hash/9b0ead00a217ea2c12e06a72eec4923f-Abstract.html}.

\bibitem[Sutton \& Barto(2018)Sutton and Barto]{sutton2018reinforcement}
Richard~S. Sutton and Andrew~G. Barto.
\newblock \emph{{Reinforcement Learning: An Introduction}}.
\newblock The MIT Press, 2nd edition, 2018.

\bibitem[Tang et~al.(2025)Tang, Chen, and Wu]{mahjongIS}
Shih-Chieh Tang, Jr-Chang Chen, and I-Chen Wu.
\newblock {Applying Importance Sampling to MCTS for Mahjong}.
\newblock \emph{IEEE Transactions on Games}, pp.\  1--10, 2025.
\newblock \doi{10.1109/TG.2025.3535740}.

\bibitem[Yoon et~al.(2007)Yoon, Fern, and Givan]{YoonFG07}
Sung~Wook Yoon, Alan Fern, and Robert Givan.
\newblock {FF-Replan: {A} Baseline for Probabilistic Planning}.
\newblock In Mark~S. Boddy, Maria Fox, and Sylvie Thi{\'{e}}baux (eds.), \emph{Proceedings of the Seventeenth International Conference on Automated Planning and Scheduling, {ICAPS} 2007, Providence, Rhode Island, USA, September 22-26, 2007}, pp.\  352. {AAAI}, 2007.
\newblock URL \url{http://www.aaai.org/Library/ICAPS/2007/icaps07-045.php}.

\bibitem[Yoon et~al.(2008)Yoon, Fern, Givan, and Kambhampati]{YoonFGK08}
Sung~Wook Yoon, Alan Fern, Robert Givan, and Subbarao Kambhampati.
\newblock {Probabilistic Planning via Determinization in Hindsight}.
\newblock In Dieter Fox and Carla~P. Gomes (eds.), \emph{Proceedings of the Twenty-Third {AAAI} Conference on Artificial Intelligence, {AAAI} 2008, Chicago, Illinois, USA, July 13-17, 2008}, pp.\  1010--1016. {AAAI} Press, 2008.
\newblock URL \url{http://www.aaai.org/Library/AAAI/2008/aaai08-160.php}.

\end{thebibliography}
\bibliographystyle{iclr2026_conference}

\appendix
\newpage
\section{Supplementary materials}
\label{sec:appendix}
\subsection{Proof of Theorem 1}
\label{sec:proof}

Running the abstraction-using UCB MCTS variant on the local-layered MDP $M_{\text{l}}$ rooted at $s_d$ and the UCT intra-abstraction policy is equivalent to running standard MCTS on a move group MDP of $M_{\text{l}}$ \citep{EyckM11} where the move groups are the abstract actions. A move group MDP splits the original MDP's actions into two phases: the first is to select a move group (which is a subset of the available actions at the corresponding state), and then select an action within that move group that has the transition dynamics of the original MDP. The move-group selection action is deterministic and yields a reward of 0. Consequently, the $Q^*$ values of the original actions in the move group MDP have the same value as in the original MDP and therefore both MDPs have identical optimal policies (when using a discount of $\gamma=1$ and excluding the move group actions). Since standard MCTS (using UCB) converges in probability to the optimal action \citep{KocsisS06} on the move-group MDP, the same must hold for the here-considered abstraction-using MCTS variant. \qed

\subsection{Intra-abstraction policy query statistics}
\label{sec:query_stats}

\begin{table}[H]
\caption{Statistics of the ratio of tree policy calls where two actions of the same parent were part of the same abstract node to show that intra-abstraction policies are not an edge case. A ratio of $1.00$ means that an intra abstraction policy was always required, while a ratio of $0.00$ means that it was never required.
    The statistics were gathered with $(0,\varepsilon_{\text{t}})$-OGA using the RANDOM intra abstraction policy, 1000 iterations, and $C=4$. The results were averaged from 100 episodes each.}
\centering
\scalebox{1.0}{
\setlength{\tabcolsep}{1mm}\begin{tabular}{c| c c|c c|c c}
\toprule
\multirow{2}{*}{Domain} & \multicolumn{2}{c|}{$\varepsilon_{\text{t}}=0$} & \multicolumn{2}{c|}{$\varepsilon_{\text{t}}=0.8$}  & \multicolumn{2}{c}{$\varepsilon_{\text{t}}=1.6$}\\
 &NO-PG & PG & NO-PG & PG & NO-PG & PG \\
\midrule

Academic Advising & 0.02 & 0.21 & 0.02 & 0.29 & 0.64 & 0.74 \\ 
Cooperative Recon & 0.30 & 0.33 & 0.33 & 0.32 & 0.28 & 0.32 \\ 
Crossing Traffic & 0.88 & 0.93 & 0.92 & 0.92 & 0.89 & 0.94 \\ 
Earth Observation & 0.00 & 0.00 & 0.27 & 0.27 & 0.27 & 0.27 \\ 
Game of Life & 0.00 & 0.00 & 0.52 & 0.84 & 0.84 & 0.82 \\ 
Manufacturer & 0.00 & 0.00 & 0.05 & 0.07 & 0.11 & 0.13 \\ 
Navigation & 0.00 & 0.00 & 0.01 & 0.02 & 0.02 & 0.13 \\ 
Racetrack & 0.17 & 0.51 & 0.17 & 0.51 & 0.17 & 0.51 \\ 
Sailing Wind & 0.00 & 0.02 & 0.00 & 0.06 & 0.05 & 0.13 \\ 
Saving & 0.00 & 0.00 & 0.00 & 0.01 & 0.01 & 0.01  \\
Skills Teaching & 0.01 & 0.03 & 0.11 & 0.11 & 0.13 & 0.20 \\ 
SysAdmin  & 0.01 & 0.05 & 0.23 & 0.30 & 0.31 & 0.38 \\ 
Tamarisk & 0.00 & 0.01 & 0.26 & 0.34 & 0.40 & 0.44 \\ 
Traffic & 0.08 & 0.13 & 0.52 & 0.77 & 0.47 & 0.79 \\ 
Triangle Tireworld & 0.32 & 0.47 & 0.28 & 0.40 & 0.33 & 0.48 \\

\bottomrule
\end{tabular}}
\label{tab:tiebreak_stats}
\end{table}

\subsection{ASAP abstraction example}
\label{sec:example_asap}

\begin{figure}[H]
    \centering
\includegraphics[width=1.0\linewidth]{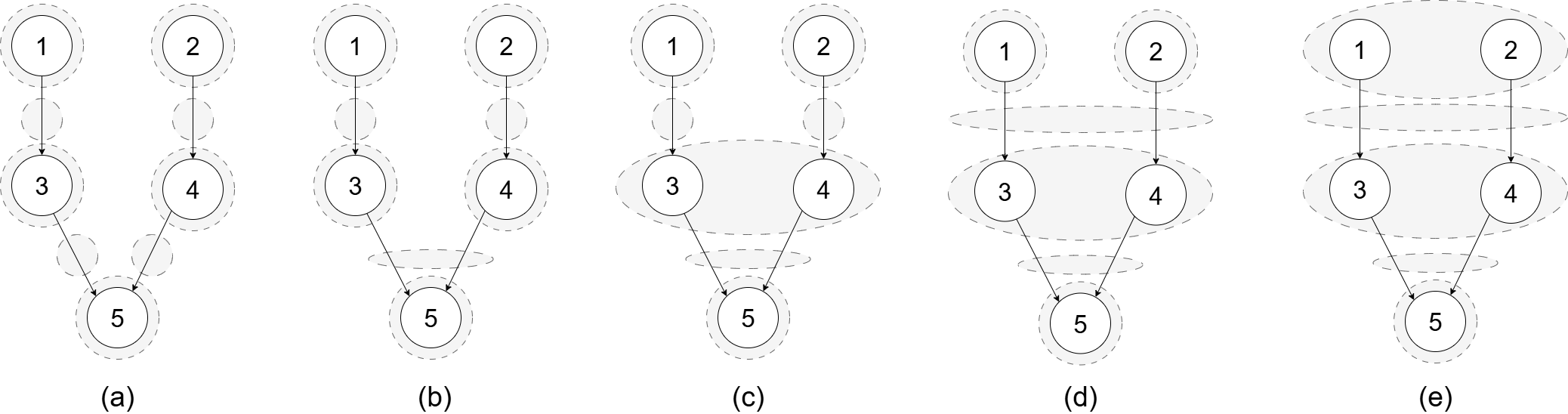}
    \caption{A showcase of how the ASAP abstraction framework, which itself is a special case of ASASAP abstractions, would detect equivalences in the following 5-state MDP. Each node represents a state, and arrows represent deterministic actions with the same immediate reward of 0. The dotted ovals represent abstractions. Initially, in (a), all states and state-action pairs are in their own singleton abstract node. Then, in (b) the next state-action pair abstraction is constructed (the application of function $f$ from Section \ref{sec:foundations}) from this initial state abstraction, which groups the actions of nodes 3 and 4 because they have the same immediate reward and the same transition distribution. From this state-action pair abstraction the next state abstraction is constructed in (c), (the application of function $g$ from Section \ref{sec:foundations}) which groups nodes 3 and 4 because they have the same set of abstract state-action pairs. Then again, in (d) the next state-action pair abstraction is constructed which also groups the actions from nodes 1 and 2 because they have the same abstract successor. Then a state abstraction is constructed again in (e), which groups states 1 and 2. Then further applications of $f$ or $g$ would have no effect, hence this abstraction is converged.}
    \label{fig:asap_example}
\end{figure}

\subsection{Performances on fixed abstractions}
\label{subsec:intra:ablation}
Next, it will be tested how well each intra-abstraction policy can generalize across different abstraction types. To do this, we reanalyzed the results of the main-part experiment section and built the pairings and relative improvement score when moving all parameters except the intra-abstraction to the set of tasks (i.e. a task now includes, for example, the $\varepsilon_{\text{t}}$ or $\alpha$ values). Tab.~\ref{tab:intra:fixedabs} shows both the relative improvement and pairings score for these results. Importantly, this shows that when confronted with an arbitrary abstraction, both FIRST and RANDOM perform worst by far, while UCT is by far the best policy.

\begin{table}[H]
\centering
\caption{The relative improvement and pairings scores for each intra-abstraction policy, when for the score calculation all parameters except the intra-abstraction policy are part of the task set.}
\label{tab:intra:fixedabs}
\scalebox{1.0}{
\setlength{\tabcolsep}{1mm}
\begin{tabular}{c c}
\toprule
Intra-abs policy &  Rel. improv. score \\
\midrule
UCT & $0.074$ \\
GREEDY & $0.046$ \\
RANDOM\_GREEDY & $0.040$ \\
LEAST\_OUTCOMES & $0.038$ \\
LEAST\_VISITS & $0.038$ \\
MOST\_VISITS & $-0.034$ \\
FIRST & $-0.100$ \\
RANDOM & $-0.101$ \\
\bottomrule
\end{tabular}
\hspace{1.5cm}
\begin{tabular}{c c}
\toprule
Intra-abs policy & Pairings score \\
\midrule
UCT & $0.394$ \\
GREEDY  & $0.206$ \\
RANDOM\_GREEDY & $0.102$ \\
LEAST\_VISITS & $0.097$ \\
LEAST\_OUTCOMES & $0.077$ \\
MOST\_VISITS & $-0.145$ \\
RANDOM & $-0.363$ \\
FIRST & $-0.368$ \\
\bottomrule
\end{tabular}
}
\end{table}

\subsection[Domain-specific $\varepsilon_{\text{a}}$ values]{Domain-specific \bm{$\varepsilon_{\text{a}}$} values}

\begin{table}[H]
\caption{A list of the environment-specific $\varepsilon_{\text{a}}$ values that were used for the experiments. All domains that are not explicitly listed here use the default values $\varepsilon_{\text{a}} \in \{0,1,2,\infty\}$. The values were chosen to be equal to reward differences (except 0 and $\infty$) that occur in these environments.}
\label{tab:ogaeps:epsa_values}
\centering
\begin{tabular}{l l}
\hline
\textbf{Environment} & $\varepsilon_{\text{a}}$ values \\
\hline
Academic Advising & 0, $\infty$ \\
Cooperative Recon & 0, 0.5, 1.0, $\infty$ \\
Crossing Traffic & 0, $\infty$ \\
Manufacturer & 0, 10, 20, $\infty$ \\
Skill Teaching & 0, 2, 3, $\infty$ \\
Tamarisk & 0, 0.5, 1.0, $\infty$ \\
Default & 0, 1, 2, $\infty$ \\
\hline
\end{tabular}
\end{table}

\subsection{Definition of the relative improvement and pairings score}
\label{subsec:scors_defs}

In the main experimental section, we evaluated the intra-abstraction policies with respect to the relative improvement and pairings scores which were first defined in \cite{aupo}.
These scores are Borda-like rankings of a set of agents which are obtained by summing over the number of games an agent outperformed another agent (pairings score) or by summing over the percentage improvements (relative improvement score). Both scores are considered in this paper since the relative improvement score alone is prone to outliers and the pairings score does not take the magnitude of the improvements into consideration.
    
The following definition of these scores is taken from \cite{aupo}.

\noindent\textbf{Definition:}
Let $\{\pi_1,\dots,\pi_n\}$ be $n$ agents (e.g., concrete parameter settings) where each agent was evaluated on $m$ tasks (e.g. a given MCTS iteration budget and an environment or a given abstraction in case of the experiments in Section \ref{subsec:intra:ablation}) where $p_{i,k} \in \mathbb{R}$ denotes the performance of agent $\pi_i$ on the $k$-th task.     The \textit{pairings score matrix} $M^{\text{pairings}} \in \mathbb{R}^{n \times n}$ is defined as 
    \begin{equation}
        M^{\text{pairings}}_{i,j} =  \frac{1}{m-1}\sum\limits_{1 \leq k \leq m}  \text{sgn}(p_{i,k}-p_{j,k})
    \end{equation}
    where sgn is the signum function.
    The \textit{pairings score} $s^{\text{pairings}}_i, i \leq n$ is given by 
    \begin{equation}
        s^{\text{pairings}}_i = \frac{1}{n-1}\sum\limits_{1 \leq l \leq n, l \neq i} M^{\text{pairings}}_{i,l}.
    \end{equation}
    The \textit{relative improvement matrix} $M^{\text{rel}} \in \mathbb{R}^{n \times n}$ is defined as 
    \begin{equation}
        M^{\text{rel}}_{i,j} = \frac{1}{m-1} \sum\limits_{1 \leq k \leq m}  \frac{p_{i,k}-p_{j,k}}{\max(|p_{i,j}|,|p_{j,k}|)}
    \end{equation}
    and the \textit{relative improvement score} $s^{\text{rel}}_i, i \leq n$ is given by 
    \begin{equation}
        s^{\text{rel}}_i = \frac{1}{n-1}\sum\limits_{1 \leq l \leq n, l \neq i} M^{\text{rel}}_{i,l}.
    \end{equation}

\subsection{RANDOM-OGA}
\label{sec:random_oga}
OGA that uses random abstractions is called RANDOM-OGA
and functions as follows. Whenever a Q node Q is visited for the K-th time and its
current abstract node consists only of itself, then with the probability $p_{\text{abs}} \in [0, 1]$, Q’s
abstract node is changed with uniform probability to any of the abstract nodes of the
same depth. Initially, at creation, any Q node is its own abstract node. Note that
RANDOM-OGA does not abstract states, as it is only state-action pair abstractions
that influence the agent’s decision-making through the modified UCB formula.

\subsection{Pairings and relative improvement scores for each individual iteration budget:}

\begin{figure}[H]
\centering

\begin{minipage}{0.95\textwidth}
\centering
\includegraphics[width=\linewidth]{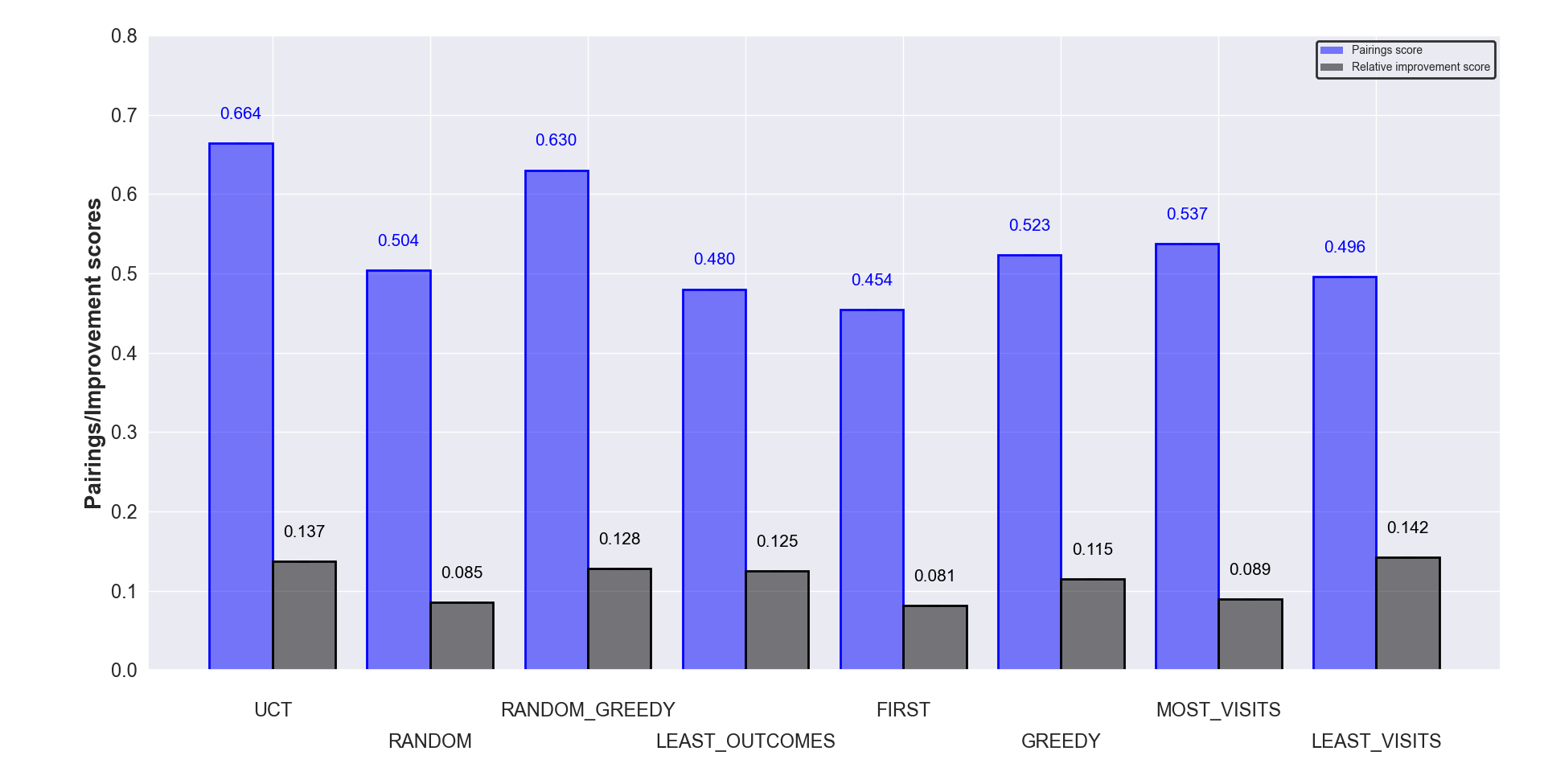}
\caption*{(a) 100 iterations}
\end{minipage}
\hfill
\begin{minipage}{0.95\textwidth}
\centering
\includegraphics[width=\linewidth]{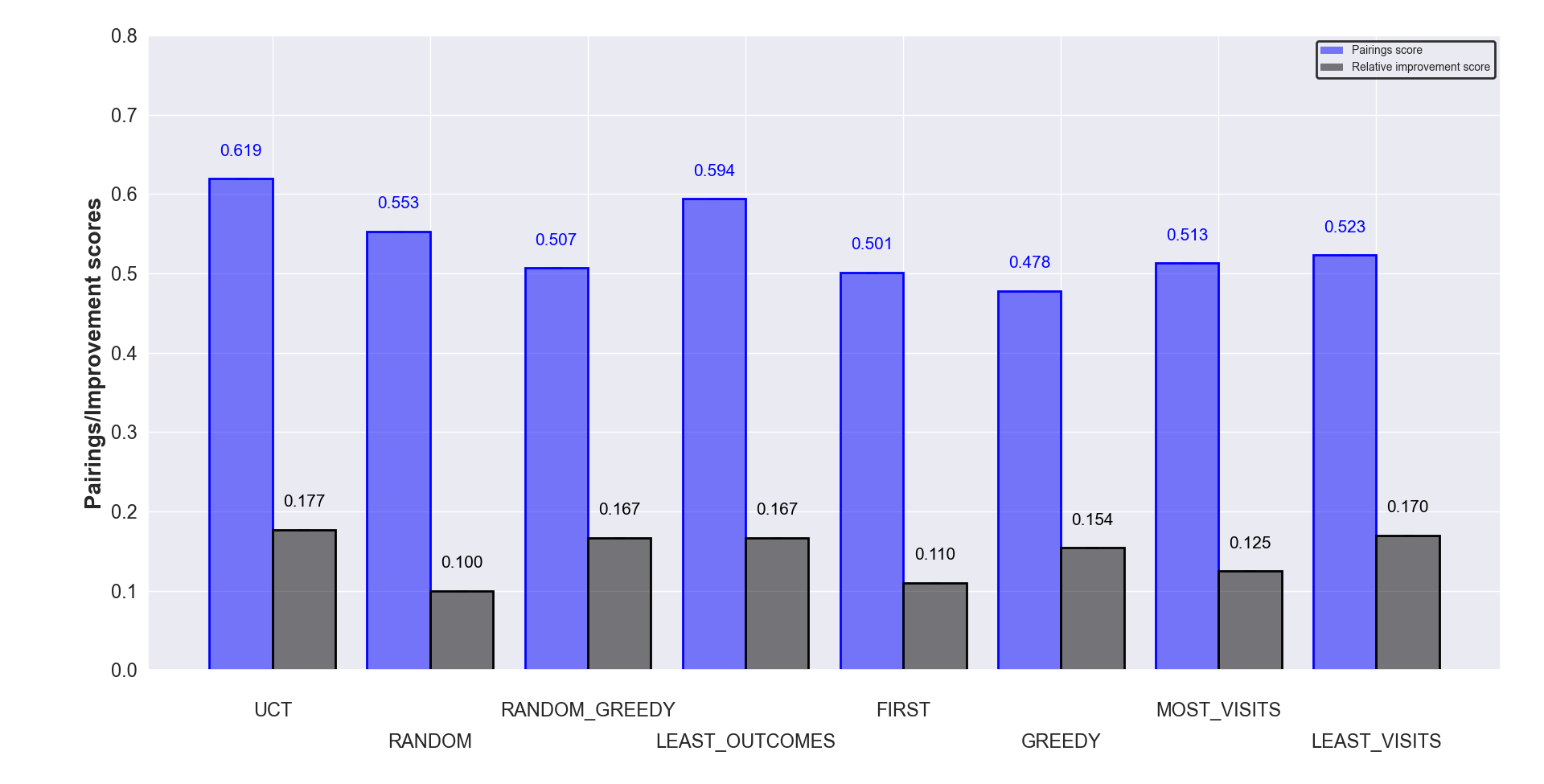}
\caption*{(b) 200 iterations}
\end{minipage}
\end{figure}

\begin{figure}[H]
\centering
\begin{minipage}{0.95\textwidth}
\centering
\includegraphics[width=\linewidth]{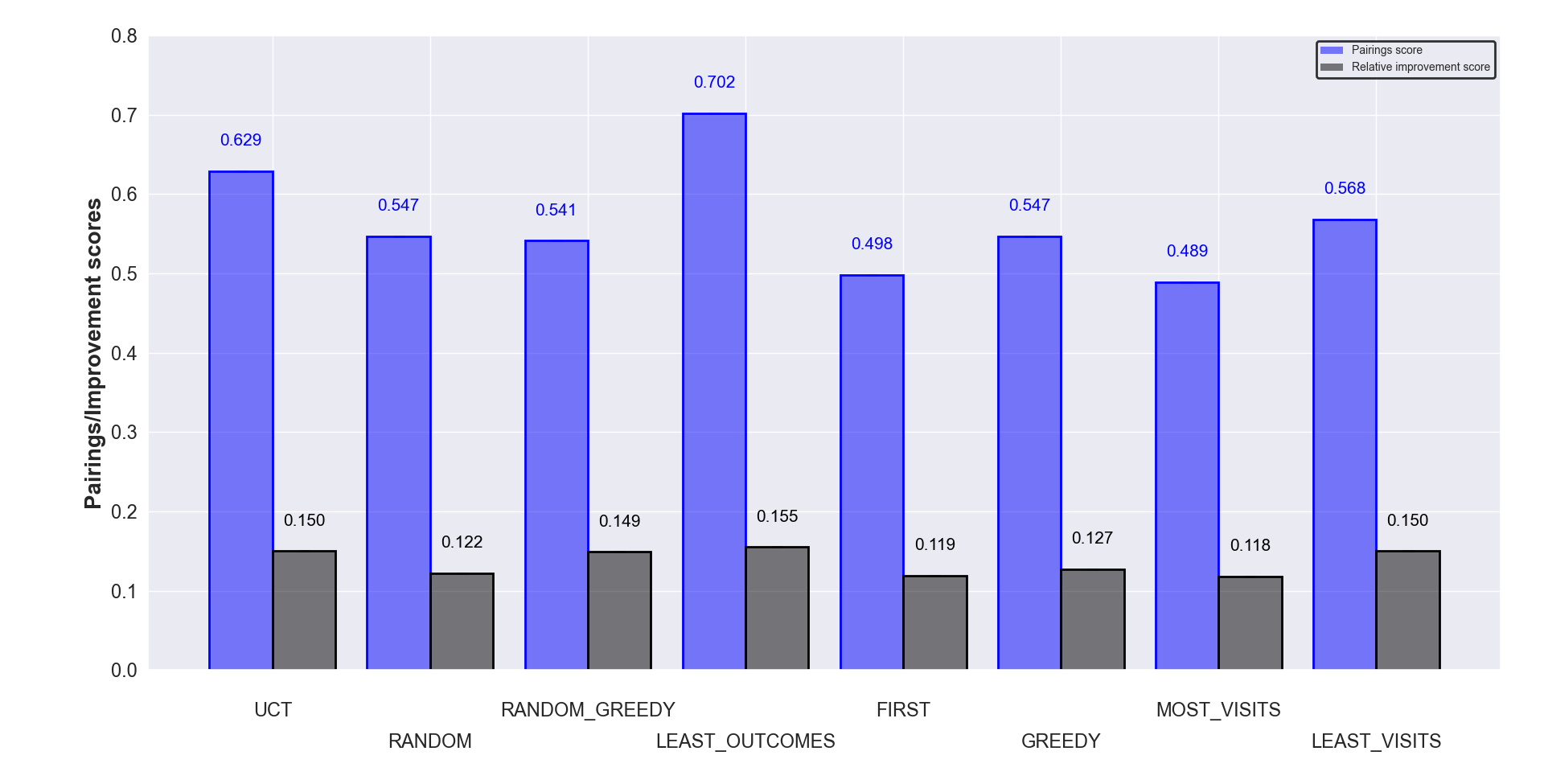}
\caption*{(c) 500 iterations}
\end{minipage}
\hfill
\begin{minipage}{0.95\textwidth}
\centering
\includegraphics[width=\linewidth]{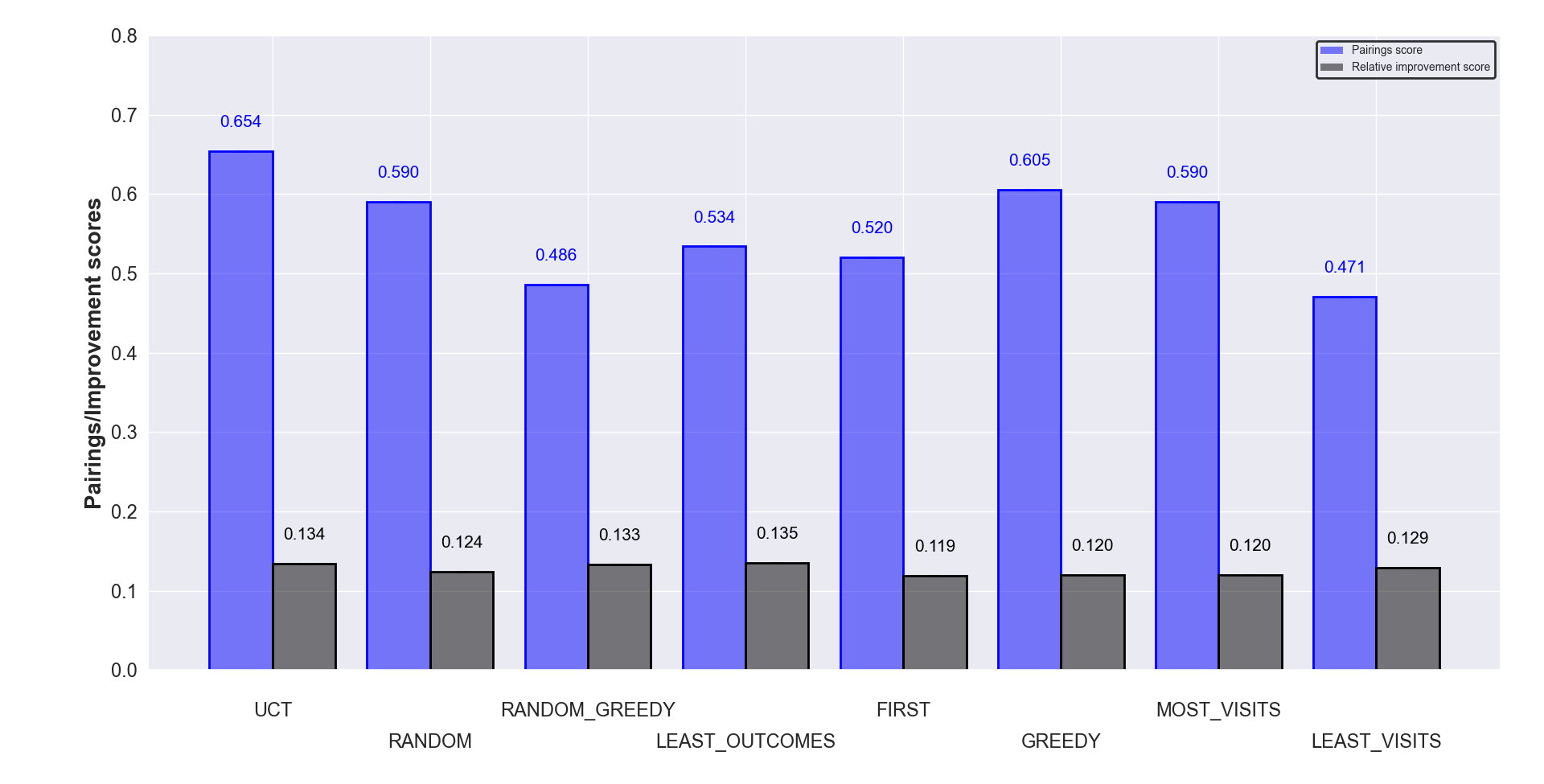}
\caption*{(d) 1000 iterations}
\end{minipage}
\hfill
\begin{minipage}{0.95\textwidth}
\centering
\includegraphics[width=\linewidth]{images/its_scores.png}
\caption*{(e) All budgets}
\end{minipage}

\caption{For each iteration budget and the combination of all iteration budgets, the pairings and relative improvement score for the best performing parameter-combination of each intra-abstraction policy are shown.}
\label{fig:intra:gen}
\end{figure}

\subsection{Ablation: Performances with varying abstraction coarsenesses}
\label{subsec:intra:coarsenesses}

\begin{figure}[H]
\centering

\begin{minipage}{0.4\textwidth}
\centering
\includegraphics[width=\linewidth]{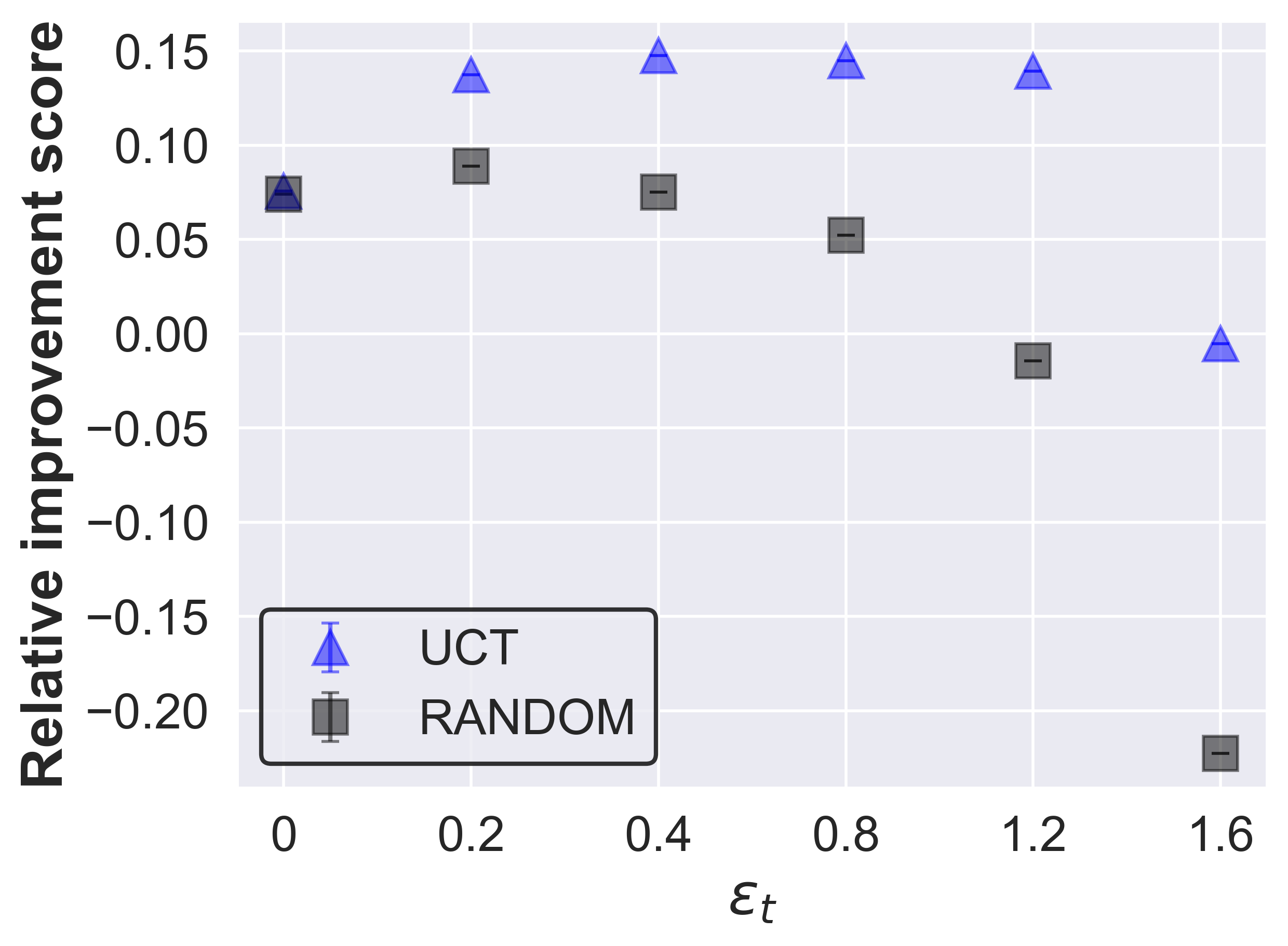}
\caption*{(a) Relative improvement score on $(\varepsilon_{\text{a}},\varepsilon_{\text{t}})$-OGA abstractions.}
\end{minipage}
\hfill
\begin{minipage}{0.4\textwidth}
\centering
\includegraphics[width=\linewidth]{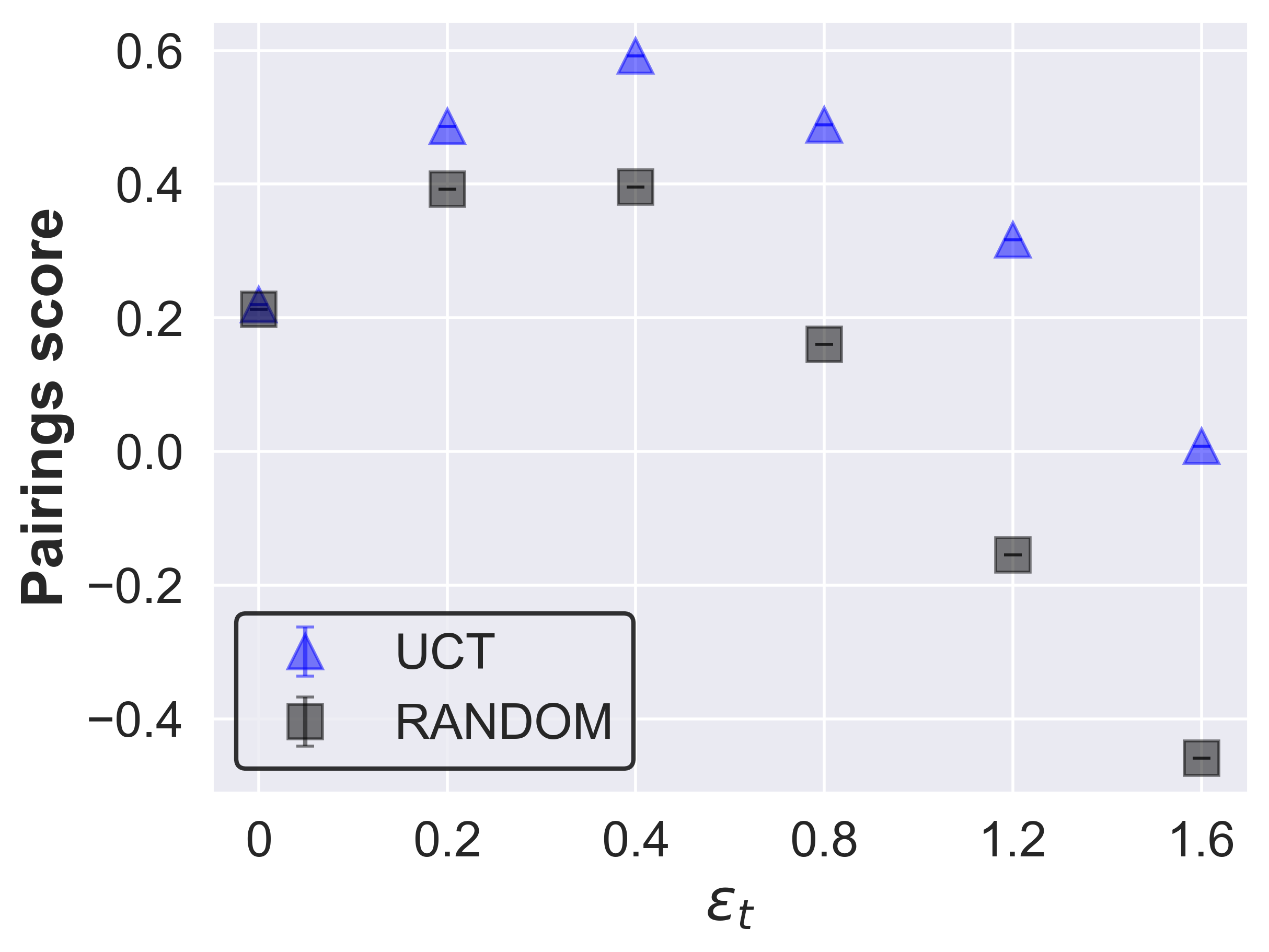}
\caption*{(b) Pairings score on $(\varepsilon_{\text{a}},\varepsilon_{\text{t}})$-OGA abstractions.}
\end{minipage}
\hfill
\begin{minipage}{0.4\textwidth}
\centering
\includegraphics[width=\linewidth]{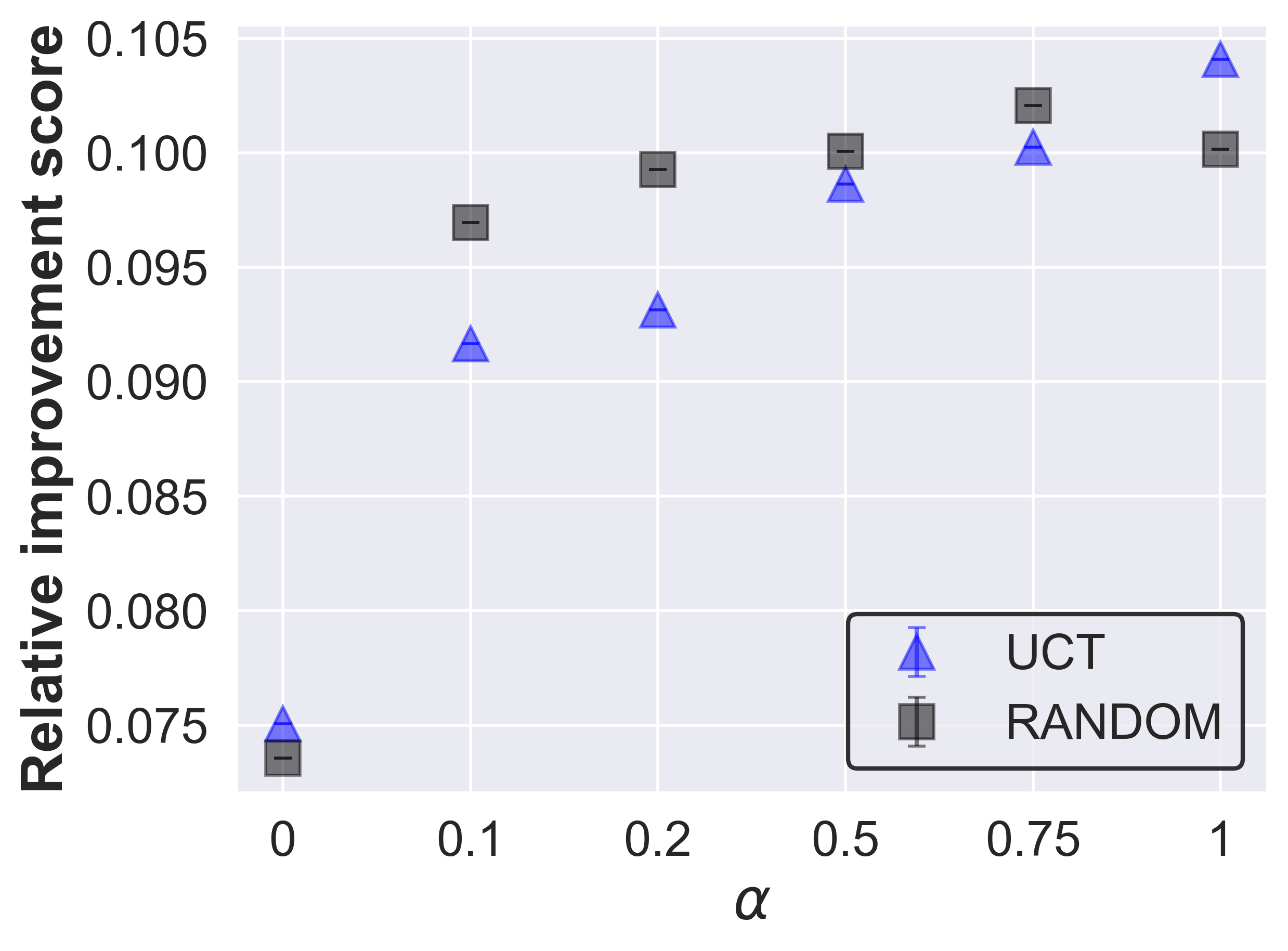}
\caption*{(c) Relative improvement score on pruned OGA abstractions.}
\end{minipage}
\hfill
\begin{minipage}{0.4\textwidth}
\centering
\includegraphics[width=\linewidth]{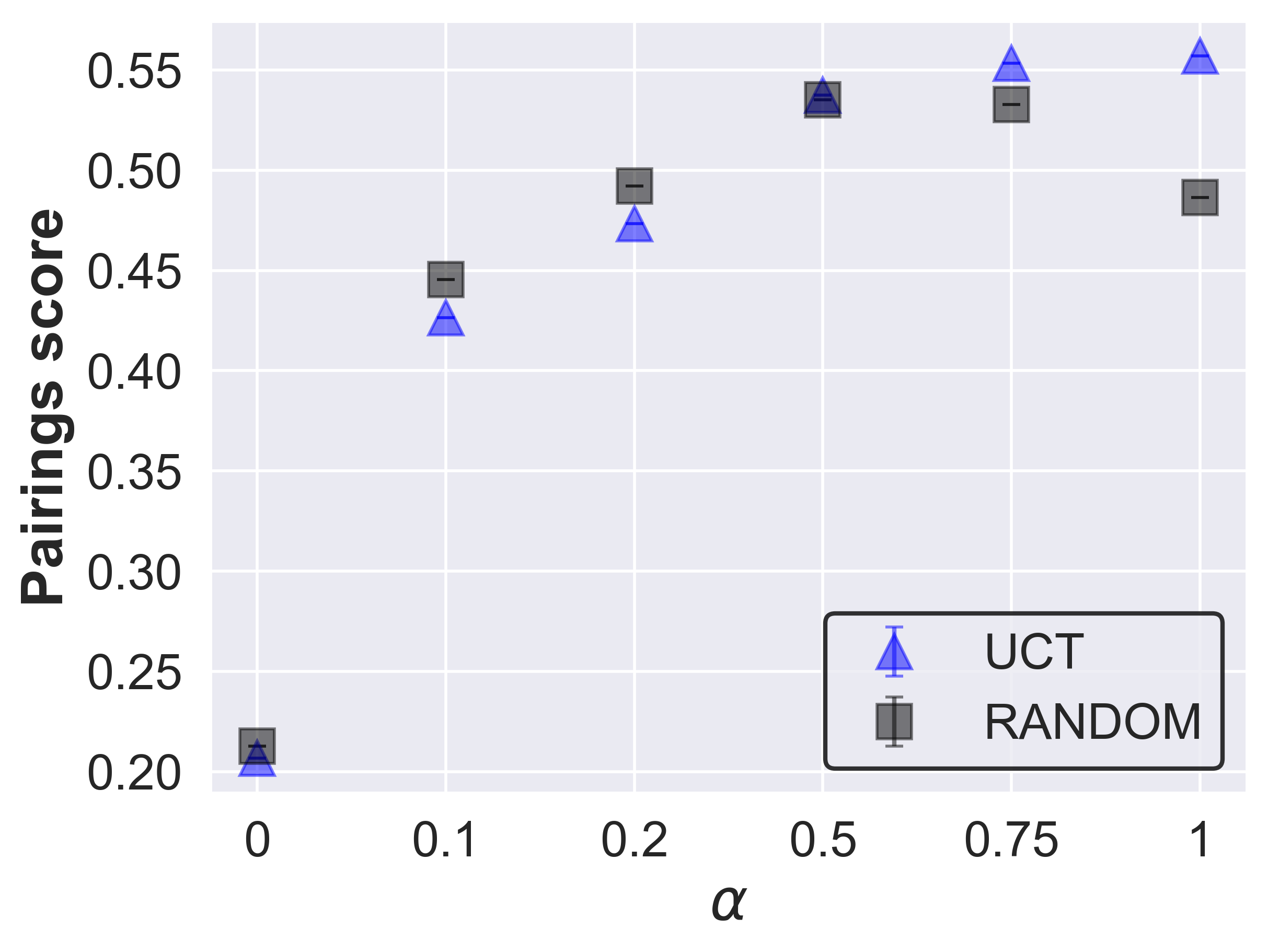}
\caption*{(d) Pairings score on pruned OGA abstractions.}
\end{minipage}
\hfill
\begin{minipage}{0.4\textwidth}
\centering
\includegraphics[width=\linewidth]{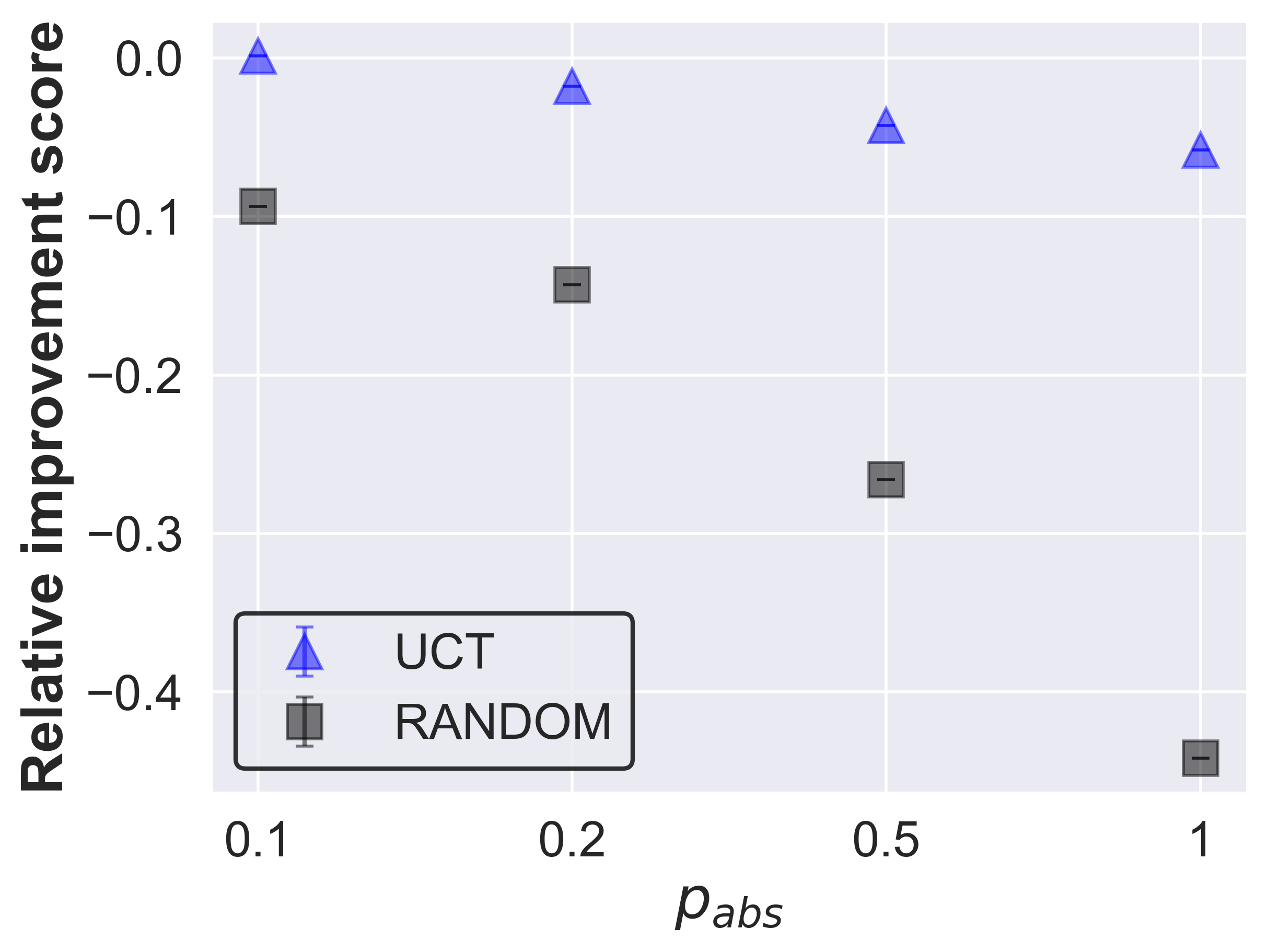}
\caption*{(f) Relative improvement score on RANDOM-OGA abstractions.}
\end{minipage}
\hfill
\begin{minipage}{0.4\textwidth}
\centering
\includegraphics[width=\linewidth]{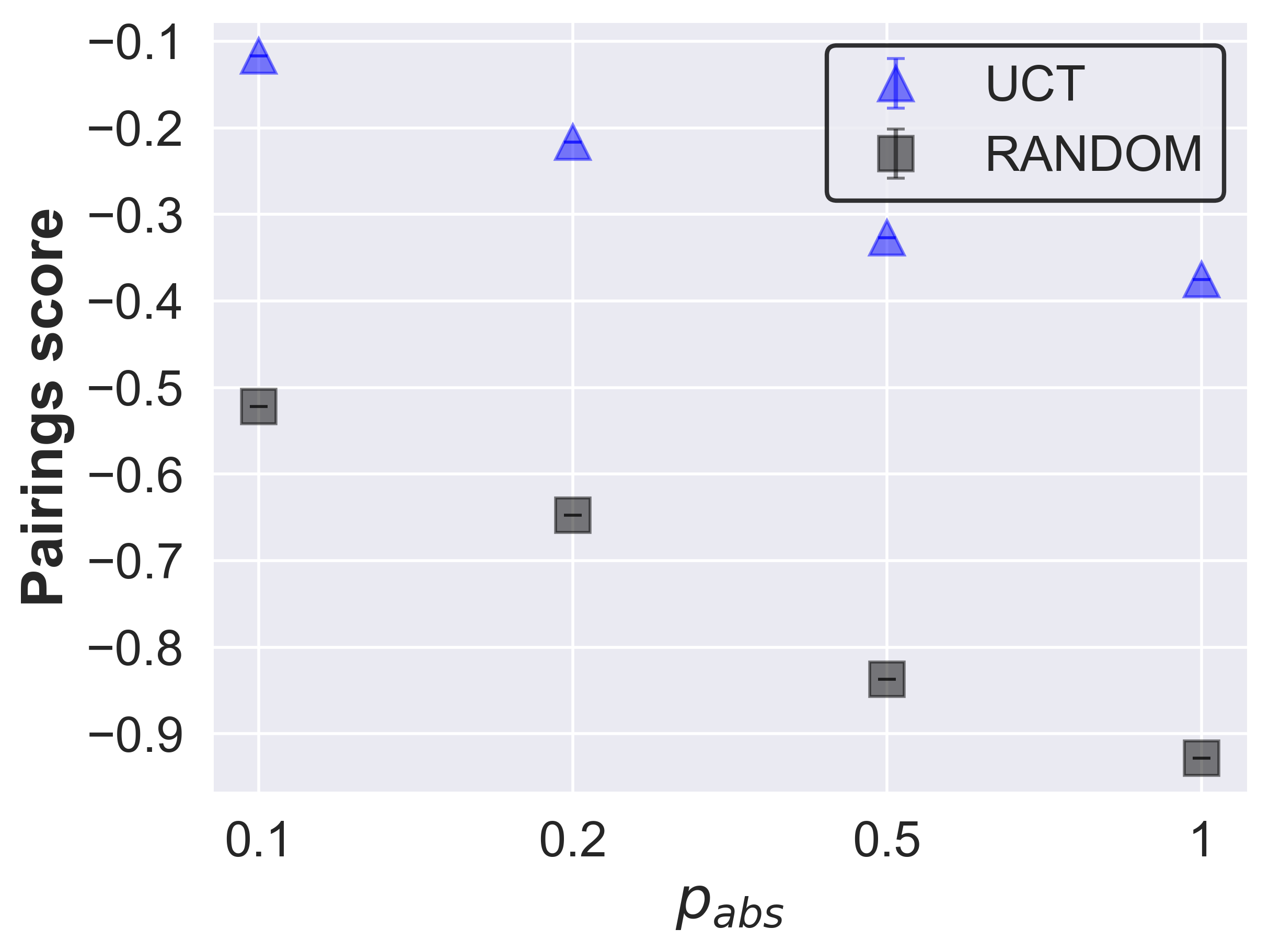}
\caption*{(e) Pairings score on RANDOM-OGA abstractions.}
\end{minipage}

\caption{For RANDOM-OGA, pruned OGA, and $(\varepsilon_{\text{a}},\varepsilon_{\text{t}})$-OGA, these figures show the parameter combinations using UCT and RANDOM as the intra-abstraction policies and the respective abstraction parameter (e.g. $\alpha=1$ that achieved the highest pairings or relative improvement score from the main-part experiment section.}
\label{fig:intra:by_coarseness}
\end{figure}

Lastly, we conducted an ablation on the efficiency of UCT and RANDOM as the intra-abstraction policy in dependence on the coarseness of the abstraction. Concretely, we reanalyzed the pairings and relative improvement scores from the experiment section that were created over all parameter combinations. Fig.~\ref{fig:intra:by_coarseness} shows the scores of several abstractions with varying coarsenesses and the highest pairings/relative improvement score that UCT and RANDOM could achieve in that setting. The results are pretty clear: The coarser the abstraction, the greater the performance gap between UCT and RANDOM. Interestingly, using UCT can change the location of the performance peaks by enabling coarser abstractions: In $(\varepsilon_{\text{a}},\varepsilon_{\text{t}})$ the peak is obtained at $\varepsilon_{\text{t}} = 0.4$ instead of $\varepsilon_{\text{t}}=0.2$, and in pruned OGA the peak is obtained at $\alpha=1$ instead of $\alpha=0.75$.

\subsection{Monte Carlo Tree Search}
\label{sec:mcts}

Next, we are going to describe the MCTS version used for the experiments of this paper.

\begin{enumerate}
    \item Instead of building a search tree, we build a directed acyclic search graph to allow for different state-action pairs in the same layer to share common successor nodes which is necessary for ASAP to detect any abstractions.
    \item For the tree policy, we use the standard UCB tree policy which selects the action at each step that maximizes the current UCB value which is defined as 
\begin{equation}
    \text{UCB}(a) = 
    \underbrace{\frac{V_a}{N_a}}_{\text{Q term}} + 
    \underbrace{\lambda \sqrt{\frac{\log\left(\sum\limits_{a^{\prime} \in \mathbb{A}(s)}N_{a^{\prime}}\right)}{N_a}}}_{\text{Exploration term}}.
\end{equation}
In this definition, $s$ is the state where the tree policy is queried, and $V_a,N_a$ is the return sum and visits of the action under consideration.
    \item For the decision policy, the root action with the maximal Q value is selected. This is the greedy decision policy.
\end{enumerate}

\subsection{Runtime measurements}
Tab.~\ref{tab:runtimes} lists the average decision-making times for each environment of the UCT intra-abstraction policy compared to RANDOM for 100 and 2000 iterations on states sampled from a distribution induced by random walks. This shows that while UCT adds only a minor overhead, despite having to execute more UCB evaluations. In particular, we are using highly optimized environment implementations that could be the runtime bottleneck in more complex environments.

\begin{table}[H]
\centering

\caption{Average decision-making times of MCTS using either the UCT or RANDOM intra-abstraction policy in milliseconds for 100 and 2000 iterations using $\varepsilon_{\text{t}} = 0.8$. This data was obtained using an Intel(R) Core(TM) i5-9600K CPU @ 3.70GHz. The data shows a median runtime overhead of $\approx$0.6\% for 100 iterations and $\approx$2\% for 2000 iterations.}
\label{tab:runtimes}
\scalebox{1.0}{
\begin{tabular}{l c c c c}
\hline
 Domain & UCT-100 & RANDOM-100 & UCT-2000 & RANDOM-2000 \\\hline

Academic Advising & 2.10 & 1.98 & 109.12 & 100.90 \\ 
Cooperative Recon & 3.67 & 3.62 & 182.63 & 179.36 \\ 
Crossing Traffic & 2.33 & 2.32 & 346.46 & 343.24 \\ 
Earth Observation & 7.38 & 7.35 & 288.69 & 290.51 \\ 
Game of Life & 3.50 & 3.54 & 149.85 & 140.11 \\ 
Manufacturer & 10.05 & 10.04 & 271.58 & 267.35 \\ 
Navigation & 2.29 & 2.20 & 61.59 & 57.69 \\ 
Sailing Wind & 2.11 & 2.12 & 144.28 & 142.40 \\ 
Saving & 1.34 & 1.37 & 70.02 & 68.03\\ 
Skills Teaching & 3.55 & 3.52 & 179.40 & 175.79 \\ 
SysAdmin  & 1.56 & 1.51 & 101.27 & 114.00 \\ 
Tamarisk & 2.81 & 3.06 & 126.52 & 117.40 \\ 
Traffic & 3.62 & 3.57 & 114.41 & 112.90 \\ 
Triangle Tireworld & 4.01 & 3.65 & 119.80 & 108.21 \\ 
\hline
\end{tabular}
}
\end{table}

\subsection{Problem descriptions}
\label{sec:problem_descriptions}

All test problems that appeared in this paper are described in \cite{mysurvey} or in \cite{demcts}.
It has to be noted that all problems can be parametrized. The concrete parameter setting can be found in the \textit{ExperimentConfigs} folder in our GitHub repository \cite{repo}.

\end{document}